\documentclass{ecai} 

\usepackage{latexsym}
\usepackage{amssymb}
\usepackage{amsmath}
\usepackage{amsthm}
\usepackage{booktabs}
\usepackage{enumitem}
\usepackage{graphicx}
\usepackage{color}

\newcommand{\BibTeX}{B\kern-.05em{\sc i\kern-.025em b}\kern-.08em\TeX}

\usepackage{algorithm}
\usepackage{algorithmic}

\DeclareMathOperator*{\E}{\mathbb{E}}

\usepackage[prependcaption,textsize=tiny]{todonotes}
\usepackage{xargs}

\newcommandx{\jsk}[2][1=]{\todo[linecolor=RoyalBlue,backgroundcolor=RoyalBlue!20,bordercolor=RoyalBlue,#1]{jsk: #2}}

\newcommandx{\erik}[2][1=]{\todo[linecolor=PineGreen,backgroundcolor=PineGreen!20,bordercolor=PineGreen,#1]{erik: #2}}

\begin{document}


\begin{frontmatter}


\paperid{M2138} 


\title{On the Effects of Irrelevant Variables in Treatment Effect Estimation with Deep Disentanglement}


\author[A]{\fnms{Ahmad Saeed}~\snm{Khan}}

\author[A]{\fnms{Erik}~\snm{Schaffernicht}}

\author[A]{\fnms{Johannes Andreas}~\snm{Stork}} 

\address[A]{Örebro University, Örebro, Sweden}


\begin{abstract}
Estimating treatment effects from observational data is paramount in healthcare, education, and economics, but current deep disentanglement-based methods to address selection bias are insufficiently handling irrelevant variables. We demonstrate in experiments that this leads to prediction errors. We disentangle pre-treatment variables with a deep embedding method and explicitly identify and represent irrelevant variables, additionally to instrumental, confounding and adjustment latent factors. To this end, we introduce a reconstruction objective and create an embedding space for irrelevant variables using an attached autoencoder. Instead of relying on serendipitous suppression of irrelevant variables as in previous deep disentanglement approaches, we explicitly force irrelevant variables into this embedding space and employ orthogonalization to prevent irrelevant information from leaking into the latent space representations of the other factors.  Our experiments with synthetic and real-world benchmark datasets show that we can better identify irrelevant variables and more precisely predict treatment effects than previous methods, while prediction quality degrades less when additional irrelevant variables are introduced.
\end{abstract}

\end{frontmatter}


\section{Introduction}
\label{sec:intro}

Treatment effect estimation from observational data is challenging because the uncontrolled mode of data collection can lead to selection bias. Selection bias causes a distributional difference between observed pre-treatment variables for different treatment groups, leading to biased counterfactual predictions. Managing this imbalance between treatment groups is therefore an important objective for improving treatment effect estimation \citep{cochran1973controlling, Johansson}.

Deep disentanglement approaches \citep{Negar,Ortho} use representation learning to identify the underlying factors as instrumental, confounding, or adjustment. This allows them to balance factors individually for improving treatment effect estimation \citep{Negar}. However, this assumes that all pre-treatment variables are pre-screened for relevance, which is impractical in increasingly prevalent data-driven and big data settings.

\begin{figure}[t]
	\centering
	\includegraphics[width=0.39\textwidth]{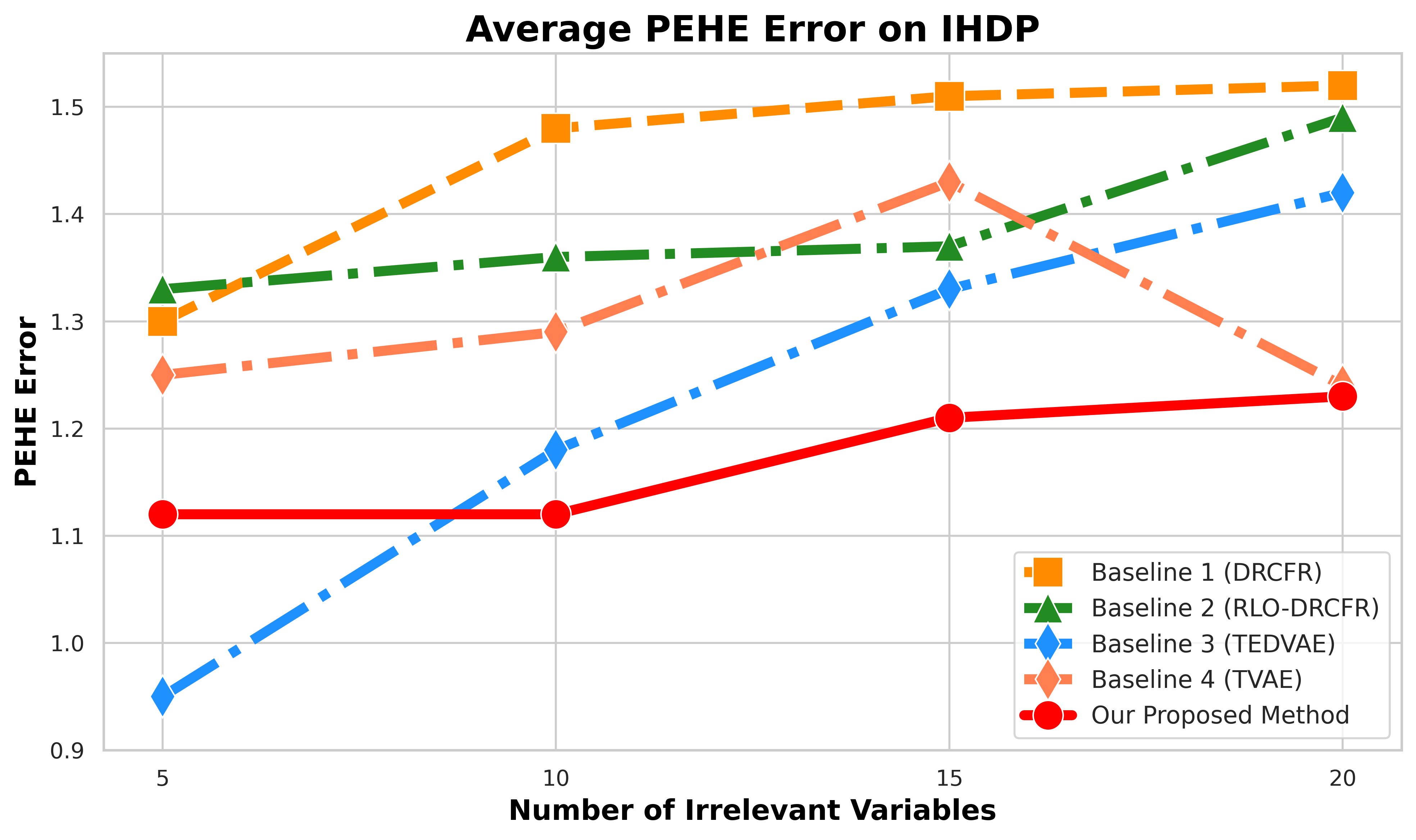}
	\caption{Average PEHE error on IHDP dataset against number of irrelevant variable dimensions (smaller the better). PEHE generally degrades with more irrelevant factors but our method is less affected.}
	\label{fig:frontpage}
\end{figure}

Our empirical analysis shows that ignoring the presence of irrelevant variables in the data critically degrades predictions with a significant drop in the precision in estimation of heterogeneous effects (PEHE) for established benchmark datasets (see Fig.~\ref{fig:frontpage}). Relying on serendipitous suppression of irrelevant variables as state-of-the-art deep disentanglement approaches do, is insufficient as it does not reliably prevent irrelevant information from leaking into other factors. Instead, it is necessary to actively disentangle irrelevant variables from other covariates used for prediction. This is supported by theoretical results emphasizing harmful consequences of unprincipled covariate inclusion \citep{pearl2010class}.

In this paper, we address the issue of unidentified irrelevant pre-treatment variables with a novel deep disentanglement approach for estimating treatment effects which explicitly identifies and represents irrelevant factors, additionally to instrumental, confounding and adjustment factors. We achieve disentanglement of irrelevant factors by introducing an additional embedding space for irrelevant factors using covariate reconstruction and orthogonality objectives.

We empirically evaluate our approach and compare it to state-of-the-art deep disentanglement baselines using the infant health and development program (IHDP), jobs and a synthetic dataset with varying number of irrelevant variables. We find that our model is better than baselines at identifying and disentangling the latent factors, including irrelevant factors, according to perturbation importance analysis \citep{breiman2001random, fisher2019all, wei2015variable} and analysis of weights of the representation networks \citep{Anpeng}. We also observe better performance on PEHE and policy risk evaluation criteria with increased number of irrelevant variables as compared to the baselines. 
Our approach is practicable and in principle compatible with previous deep disentanglement-based works as the additional channel and reconstruction objective leave the other representation networks and objectives unaltered.

Our core contributions are:
\vspace{-0.2cm}
\begin{itemize}
	\item Investigating the impact of irrelevant variables on estimation of treatment effects for state-of-the-art disentangled representational learning methods.
	
	\item The proposal of an autoencoder-based approach to disentangle irrelevant factors explicitly. 
	
	\item A thorough evaluation of our approach, showing that it outperforms the baseline methods in estimating individual treatment effects, especially in the presence of irrelevant variables.
\end{itemize}

\section{Related Work}  
\label{sec:related-work}


	
	Selection bias in observational data is a well known problem in treatment effect estimation, which is classically countered by balancing confounders with matching, stratification, and re-weighting approaches \citep{PaulRD, PaulR, Sheng}. However, assuming unfoundedness or relying on prior knowledge of the causal structure is unsuitable in real-world, high-dimensional settings, leading to underperformance \citep{conf}.
	
	
	
	Deep representation learning has been proposed for balancing variables in higher-dimensional settings \citep{Johansson, SITE, UriSha, Negar_2}.  The idea behind these approaches is to make the embedded data look like a Randomized Controlled Trial (RCT) by minimizing the discrepancy between treatment groups. 
	Maximum mean discrepancy \citep{gretton2009covariate} and Wasserstein distance \citep{villani2009optimal} are integral probability measures that are both used as discrepancy losses in these methods. We also employ the latter to balance our embedding spaces. The appealing simplicity of balancing all covariates in a shared single embedding space, however, overlooks the fact that usually not all covariates contribute to both, treatment and effect. 
	
	
	Disentanglement approaches, in contrast, account for the underlying causal structure by creating separate representations for instrumental, confounding and adjustment factors \citep{Negar, Decomp, kuang2020data, Anpeng, Ortho}. Disentangled representations give insights and can be used to reduce, as well as account for, the negative impact of selection bias \citep{Negar}. A major challenge for these approaches is imperfect decomposition with pre-treatment variables information leaking into unrelated factors, which can degrade performance on the downstream prediction task. 
	To ensure better separation, several techniques have been proposed such as different orthogonalization \citep{Decomp, kuang2020data, Anpeng} and mutual information \citep{Ortho} objectives. While \citet{Decomp, kuang2020data} only consider linear embeddings, \citet{Anpeng} present a deep orthogonal regularizer for deep representation networks, which we also use in this work. \citet{TEDEV} propose to use variational autoencoders (VAE) for separating the three factors (TEDVAE). VAEs have been used previously in \citet{CEVAE} to estimate the confounding factors only, without the use of any discrepancy loss. \citet{wu2021beta} present a prognostic score-based VAE approach to estimate causal effects for data with limited overlap between treatment groups. 
	
	
	
	
	
	In real-world observational studies unidentified, irrelevant variables are inevitable and our empirical results show that irrelevant variables can degrade prediction results. Removing irrelevant variables has been studied in feature selection field \citep{guyon2003introduction} and many representation learning approaches are implicitly considered to remove irrelevant information. Yet there are indications \citep{Leo, Kim} that this is not sufficient, especially for tabular data. However, applying classic feature selection in a disentanglement task for treatment effect estimation is not straight forward due to proxy variables. 
	
	Some of the disentanglement-based works discussed before consider irrelevant variables: \citet{Negar} includes one single Gaussian noise factor in their synthetic dataset for evaluations; \citet{Anpeng} claim that orthogonal regularization reduces the influence of irrelevant variables on the prediction; and \citet{Decomp} mention that they eliminate irrelevant variables in their linear embedding with $L_1$ penalties. While the latter analyze separation of confounders and adjustment factors, they do not report on the identification of irrelevant factors. In contrast to our work, none of the approaches above explicitly represents irrelevant factors to achieve disentanglement. All of the approaches above rely on serendipitous suppression of irrelevant factors which is in some cases encouraged by regularization.
	
	\begin{figure}
		\centering
		\includegraphics[width=0.42\textwidth]{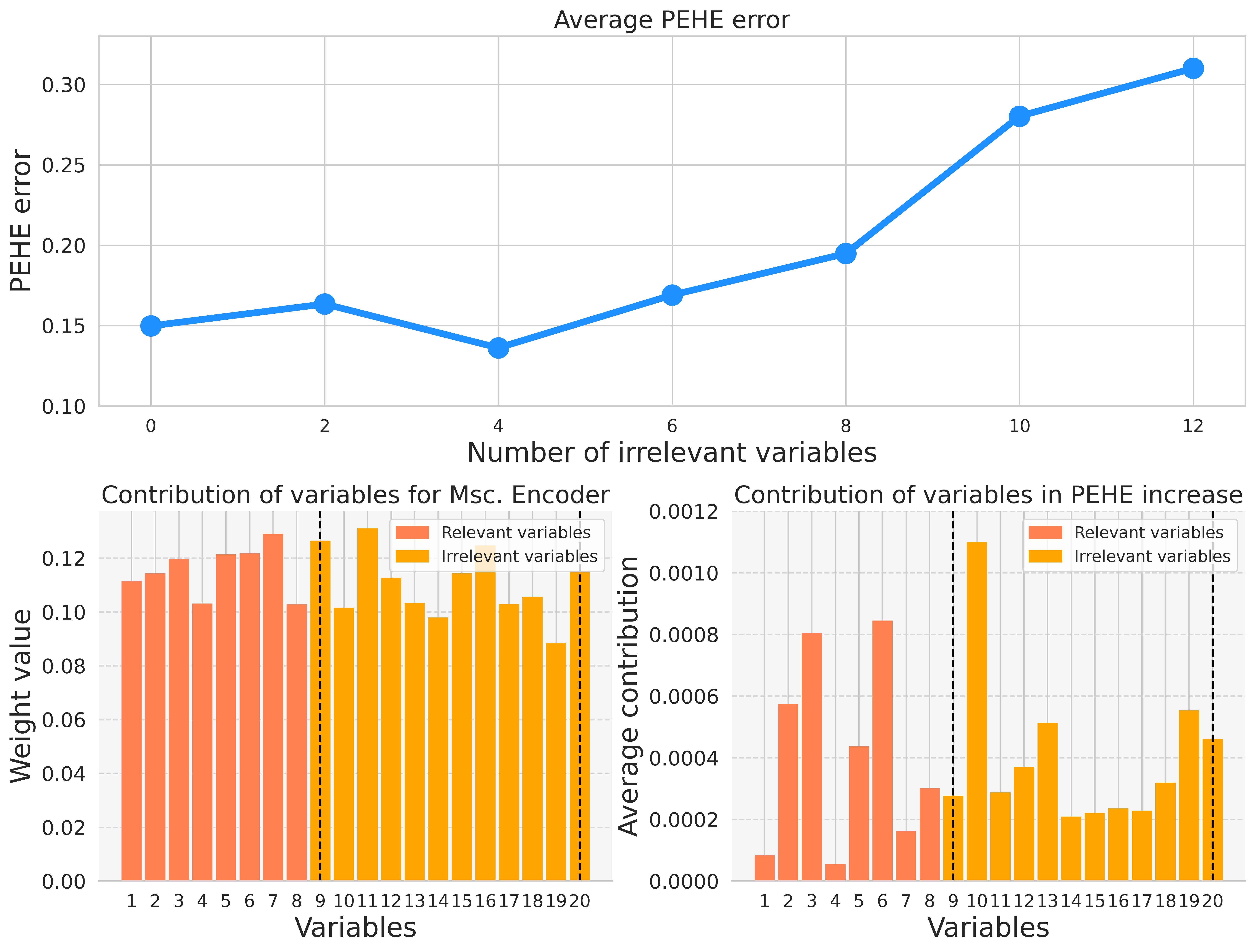}

		
		\caption{(Top): Illustrates the rise in PEHE as the number of irrelevant variables grows based on a baseline approach (TVAE). (Left): visualizes the individual contributions of variables towards learning the encoder for miscellaneous factors. Notably, the contribution of irrelevant variables mirrors that of relevant ones, underscoring the limitations of TVAE in disentangling irrelevant variables. (Right): shows average contribution of each variable in PEHE increase by using permutation of variables. Irrelevant variables are significantly participating in PEHE increase. }\label{fig:targeted_VAE}
	\end{figure}

	 Targeted VAE (TVAE) is the most similar work to our approach. It employs an encoder to manage miscellaneous factors \citep{vowels2021targeted}, but a crucial distinction arises. While TVAE aims to disentangle irrelevance when it's intertwined with relevant variables, it falls short in identifying and disentangling irrelevant variables existing in separate dimensions in pre-treatment variables. Our findings, depicted in Figure \ref{fig:targeted_VAE}, reveal a notable increase in PEHE error as irrelevant variables increase based on TVAE. Furthermore, the encoder designed for miscellaneous factors demonstrates an inability to disentangle these variables within the data. Additionally, as depicted in Figure \ref{fig:targeted_VAE}, it becomes evident that TVAE faces challenges in mitigating the influence of irrelevant variables, which contribute to the increase in PEHE. These observations are based on the same synthetic data utilized in the original study by \citet{vowels2021targeted}.

	
	
	
	

\section{Formalization and Assumptions} 
\label{sec:Notations}


In this section, we first give the notations and assumptions for treatment effect estimation in observational data. Moreover, we also define underlying latent factors of pre-treatment variables.

Formally, observational studies have a dataset: $\mathcal{D}=\{x_i, t_i, y_i\}_{i=1}^N$, where the $i^{th}$ instance has some contextual information ${x_i} \in \mathcal{X} \subseteq \mathbb{R}^K$ (often called pre-treatment variables: e.g., gender and age), ${t_i} $ is the observed treatment from the set of treatments $\mathcal{T}$ (e.g., {0: medication, 1: surgery}) and ${y_i} \in \mathcal{Y}$ (e.g., recovery time; $\mathcal{Y}\subseteq \mathbb{R}^+$)  is the respective outcome as the result of particular treatment ${t_i} $. In data $\mathcal{D}$, we only observe one outcome against the used treatment (known as factual outcome $ y^{t}_{i}$) but alternative output (counterfactual outcome $y^{\neg t}_{i}$) is never observed in the data.
In such datasets, $\mathcal{X}$ influences treatment assignment policy which causes selection bias in the data, where the condition $ P(\mathcal{T}|\mathcal{X}) = P(\mathcal{T}) $ does not hold and it lacks RCT properties\citep{Imbens,Negar}. 

\begin{figure}[t]
	\centering
	\includegraphics[width=0.40\textwidth]{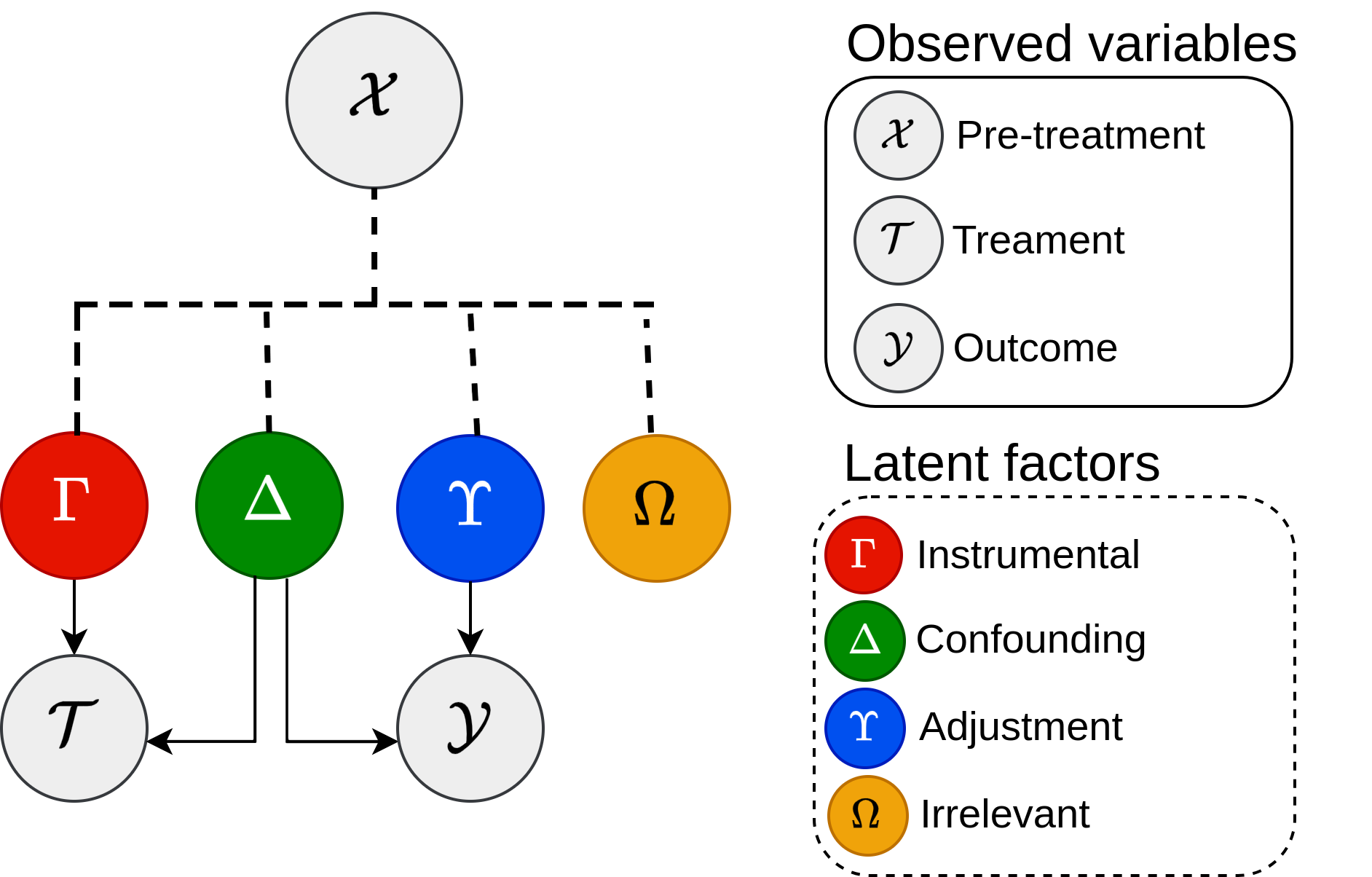}
	\caption{Underlying factors of $\mathcal{X}$. Observe that $\Omega$ has no associated downstream task with any observed variable.}
	\label{fig:latent}
\end{figure}


\citet{Negar} assume, without loss of generality, that $\mathcal{X}$ is generated by unknown joint distribution $P(\mathcal{X} \mid \Gamma,\Delta,\Upsilon)$, where $\Gamma,\Delta, \Upsilon$ are latent factors; we are keeping the previously established notations for the these factors. $\Gamma$ (instrumental factors) only influence treatment selection, $\Delta$ (confounding factors) affect both treatment selection and outcome, while $\Upsilon$ (adjustment factors) impact outcome only. We assume that there is another underlying irrelevant latent factor ($\Omega$) behind the generation of $\mathcal{X}$, depicted in Figure \ref{fig:latent}. Moreover, we also assume that latent factors are associated with separate dimensions of $\mathcal{X}$ as stated in \citet{Decomp}. Learning the representation of $\Omega$ helps to match the true data generation process without harming the identifiability of causal effects \citep{vowels2021targeted}.   


The objective of this paper is to estimate Individual Treatment Effect (ITE) for each $x_{i}$: $ \mathit{ite}_{i} = y^{1}_{i} - y^{0}_{i}$, by learning a function $f \colon \mathcal{X} \times \mathcal{T} \rightarrow \mathcal{Y}$. However, it is not straightforward to learn such function $f$ because $\mathcal{D}$ contains selection bias and irrelevant factors ($\Omega$). It is essential to disentangle $\Omega$ from other latent factors to efficiently mitigate selection bias and to have reliable estimate of ITE by avoiding the overfitting of regression function $f$ \citep{Decomp}. Empirically, we have observed a decline in the performance of recent disentangled representation learning methods for the ITE estimation with the increasing presence of $\Omega$, as shown in the Figure \ref{fig:frontpage}.


Our work, like other methods in this domain, also relies on three assumptions as presented in \citet{Rubin}. 


\begin{enumerate}
	\item \textbf{Stable Unit Treatment Value:} The treatment assignment to one unit does not affect the distribution of potential outcomes of the other unit.
	\item \textbf{Unconfoundedness:} There is no unmeasured confounding. All confounding effect on $\mathcal{Y}$ and $\mathcal{T}$ has been measured, formally, $\mathcal{Y} \perp\!\!\!\perp \mathcal{T} \mid \mathcal{X} $.
	
	\item \textbf{Overlap:} assumption states that the probability of assigning any treatment to $x$ is higher than zero. Formally, $P(t \mid x) > 0  \, \forall t \in \mathcal{T}, \forall x \in \mathcal{X} $.
	
\end{enumerate}

The assumptions of unconfoundedness and overlap are jointly known as strong ignorability.






\section{Methods}
\label{sec:method}




Considering the likelihood of $\Omega$ being present in observational data, it becomes imperative to devise an approach that disentangles $\Omega$ and estimates ITE robustly.

\begin{figure}[h]
	\centering
	\includegraphics[width=0.4\textwidth]{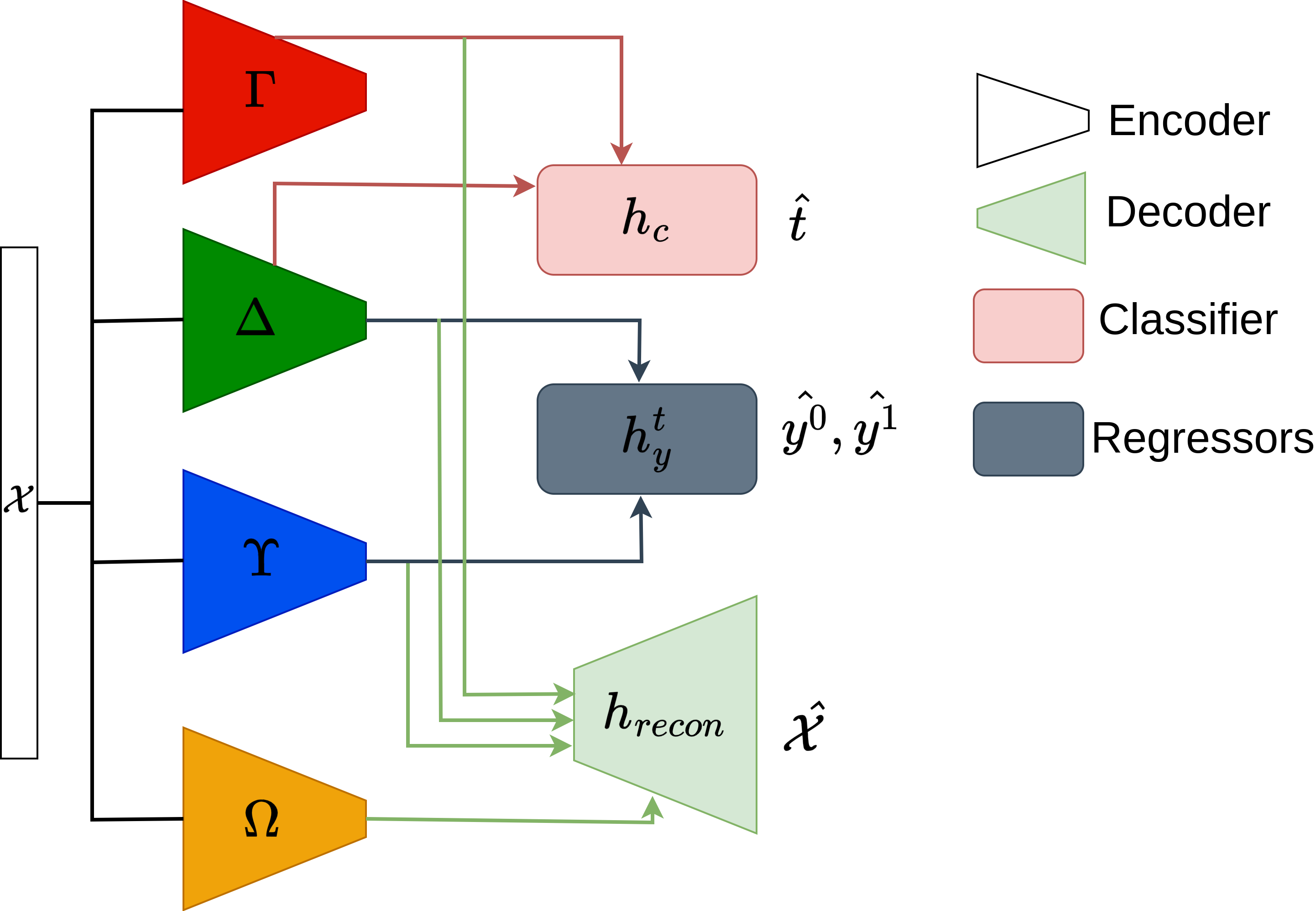}
	
	\caption{High level architecture of DRI-ITE.}
	\label{fig:archi}
\end{figure}


Thereto, we propose Disentangled Representation with Irrelevant Factor for Individual Treatment Effect Estimation (DRI-ITE) for the binary treatment case, which learns disentangled representation with four latent factors ($\Gamma$,  
$\Delta$, $\Upsilon$, $\Omega$), accounting for selection bias and simultaneously learns to predict counterfactual outcome for the final estimate of treatment effect. We achieve disentanglement of $\Omega$ by introducing an additional embedding space for irrelevant factors using $\mathcal{X}$ reconstruction and orthogonality objectives.


Figure \ref{fig:archi} shows the DRI-ITE architecture, which contains four representational networks (encoders). Each network learns one specific latent factor. Two regression networks (one for each treatment group) learn to predict factual and counterfactual outcomes, and help two representational networks to disentangle $\Delta$ and $\Upsilon$ using $\mathcal{L}_{reg}$. One classification network learns to predict the treatment and helps in disentangling $\Gamma$ and $\Delta$ using $\mathcal{L}_{class}$. 

Finally, estimating $\Omega$ directly is difficult since there is no associated downstream task as shown in Fig.\ref{fig:latent}. Instead, we employ a decoder that reconstructs $\mathcal{X}$. The core idea is that reconstructing the input data $\mathcal{X}$ in this autoencoder fashion requires to capture all latent factors including those not relevant to the ITE estimation. Intuitively, this allows us to use orthogonality objectives to separate the irrelevant factors into their own embedding space as $\Omega = \mathcal{X} \setminus \{ \Gamma, \Delta,\Upsilon$\}. 

From a computational point of view, DRI-ITE is moderately more expensive than comparable approaches due to the extra embedding space to learn the irrelevant factors.

The formal algorithm is provided in \ref{alg:algo1} and we will discuss details of each loss function in the following.

\begin{algorithm}[] 
	\caption{Disentangled Representation with Irrelevant Factor for Individual Treatment Effect Estimation (DRI-ITE)}
	\label{alg:algo1}
	\textbf{Input}:{ $\mathcal{D}= \{x_1, t_1, y_1\},...,\{x_N, t_N, y_N\}$}\\
	\textbf{Output:} $\hat{y}^{1},\hat{y}^{0}$\\
	\textbf{Loss function:} $\mathcal{L}_{main}$\\
	\textbf{Components:} \raggedright Four representation networks $\{\Gamma(.),\Delta(.),\Upsilon(.),\Omega(.)\}$, two regression networks \{ $h_{y}^{0}(.),h_{y}^{1}(.)$\}, one decoder $ h_{recon}(.)$ and one classification network $h_{c}$ \par
	\begin{algorithmic}[1] 
		
		\STATE \textbf{for} {$i$ = 1 \TO N }
		\STATE $\{x_i, t_i, y_i\}_{i=1}^N \rightarrow \{ \Gamma(x_{i}),\Delta(x_{i}),\Upsilon(x_{i}),\Omega(x_{i})\}$ 
		\STATE $h_{c}(\Gamma(x_{i}),\Delta(x_{i})) \rightarrow \hat{t_{i}}$
		\STATE $h_{y}^{0}(\Delta(x_{i}),\Upsilon(x_{i})),h_{y}^{1}(\Delta(x_{i}),\Upsilon(x_{i})) \rightarrow$ $\hat{y}^{1},\hat{y}^{0}$\\
		\STATE $h_{recon}(\Gamma(x_{i}),\Delta(x_{i}),\Upsilon(x_{i}),\Omega(x_{i})) \rightarrow x_{i}$
		\STATE $ w \leftarrow$ Adam$\{\mathcal{L}_{main} \}$
		\STATE \textbf{end for}
		\STATE \textbf{return} $\hat{y}^{1},\hat{y}^{0}$
		
	\end{algorithmic}
	
\end{algorithm}
The main objective function to be minimized is as follows: 
\begin{eqnarray}
	\begin{aligned}
		\mathcal{L}_{main}= & \mathcal{L}_{reg}+\alpha\cdot\mathcal{L}_{class}+\beta\cdot\mathcal{L}_{disc} \\ & +\gamma\cdot\mathcal{L}_{recons}+\lambda\cdot\mathcal{L}_{orth} \\ 
		& +\mu\cdot Reg(h_{y}^1,h_{y}^0,h_{c},h_{recon}).
	\end{aligned}
\end{eqnarray}
$Reg$ is a regularization term for the respective functions and $\alpha,\beta,\gamma,\lambda,\mu$ are weighting parameters. 

We define $\mathcal{L}_{reg}$ as:
\begin{eqnarray}
	\mathcal{L}_{reg}=\mathcal{L}[y_{i},h^{ti}_{y}(\Delta(x_{i}),\Upsilon(x_{i}))].
\end{eqnarray}
\textbf{{$\mathcal{L}_{reg}$}} is the regression loss (Mean squared error: MSE). We train two regression networks as used in \citet{UriSha} and \citet{Negar} to predict observed outcome based on respective treatment.  It is noteworthy that these regressors are learning on the concatenation of the $\Delta$ and $\Upsilon$ factors. Minimizing $\mathcal{L}_{reg}$ ensures that information regarding the outcome $y$ is retained in these two latent factors and both representational networks learn its respective factors.

We define $\mathcal{L}_{class}$ as:
\begin{eqnarray}
	\mathcal{L}_{class}=\mathcal{L}[t_{i},h_{c}(\Gamma(x_{i}),\Delta(x_{i}))].
\end{eqnarray} 
\textbf{{$\mathcal{L}_{class}$}} is the classification loss (Binary cross-entropy: BCE). Classifier $h_{c}$ learns to predict the treatment using the concatenation of $\Delta$ and $\Gamma$.

We define $\mathcal{L}_{disc}$ as follows:
\begin{eqnarray}
	\mathcal{L}_{disc}=disc[\Upsilon(x_{i})t_{i=0},\Upsilon(x_{i})t_{i=1}].
\end{eqnarray}
By minimizing \textbf{{$\mathcal{L}_{disc}$}}, we ensure that $\Upsilon$ contains no influence from $\Gamma$. In other words, \textbf{{$\mathcal{L}_{disc}$}} helps to mitigate selection bias caused by $\Gamma$ to have unbiased predictions for the downstream task. We use the Wasserstein distance as discrepancy loss as proposed by \citet{Ortho}.


The definition of $\mathcal{L}_{recons}$ is as follows:
\begin{eqnarray}
	\mathcal{L}_{recons}=\mathcal{L}[x_{i},h_{recon}(\Gamma(x_{i}),\Delta(x_{i}),\Upsilon(x_{i}),\Omega(x_{i}))].
\end{eqnarray}
\textbf{{$\mathcal{L}_{recons}$}} is the reconstruction loss (MSE) used by the autoencoder to reconstruct $\mathcal{X}$ based on all four embedding spaces.

$\mathcal{L}_{orth}$ is deep orthogonal regularizer to ensure distinction among latent factors. Its idea is originally inspired by \citet{Decomp}. We used the loss $\mathcal{L}_{orth}$ in the same way as it was used by \citet{Anpeng}. However, instead of constraining orthogonality on pairs of average weight vectors of just three representational networks, we constrain orthogonality for the three more pairs to keep $\Omega$ separate from all other three basic factors. We define $\mathcal{L}_{orth}$ as follows:
\begin{eqnarray}
	\begin{aligned}
		\mathcal{L}_{orth}= & \bar{W}_{\Gamma}^{T}\cdot\bar{W}_{\Delta}+\bar{W}_{\Delta}^{T}\cdot\bar{W}_{\Upsilon}+\bar{W}_{\Upsilon}^{T}\cdot\bar{W}_{\Gamma} \\ & +\bar{W}_{\Omega}^{T}\cdot\bar{W}_{\Gamma}+\bar{W}_{\Omega}^{T}\cdot\bar{W}_{\Delta}+\bar{W}_{\Omega}^{T}\cdot\bar{W}_{\Upsilon},
	\end{aligned}
\end{eqnarray}
where $W\subseteq \mathbb{R}^{d\times d}$ is the product of weight matrices across all layers within a representational network, $\bar{W}\subseteq \mathbb{R}^{d \times 1}$ represents the row-wise average vector of the absolute values of $W$ for each network. The vector $\bar{W}$ provides insight into the average contribution of each feature within that specific representational network. When $\mathcal{L}_{orth}$ is minimized, the dot products between the weight vectors become small or close to zero indicating orthogonality. Orthogonality between representations encourages each representational network to focus on capturing unique patterns and features relevant to its specific task. It prevents the networks from redundantly learning similar information. Alternatively, concepts from information theory i.e. total correlation or mutual information can also be utilized to separate information between the representational networks, but given the computational constraints of these methods it is common to employ deep orthogonal regularizers.  



\begin{figure*}
		\centering
		
			\includegraphics[width=0.22\textwidth]{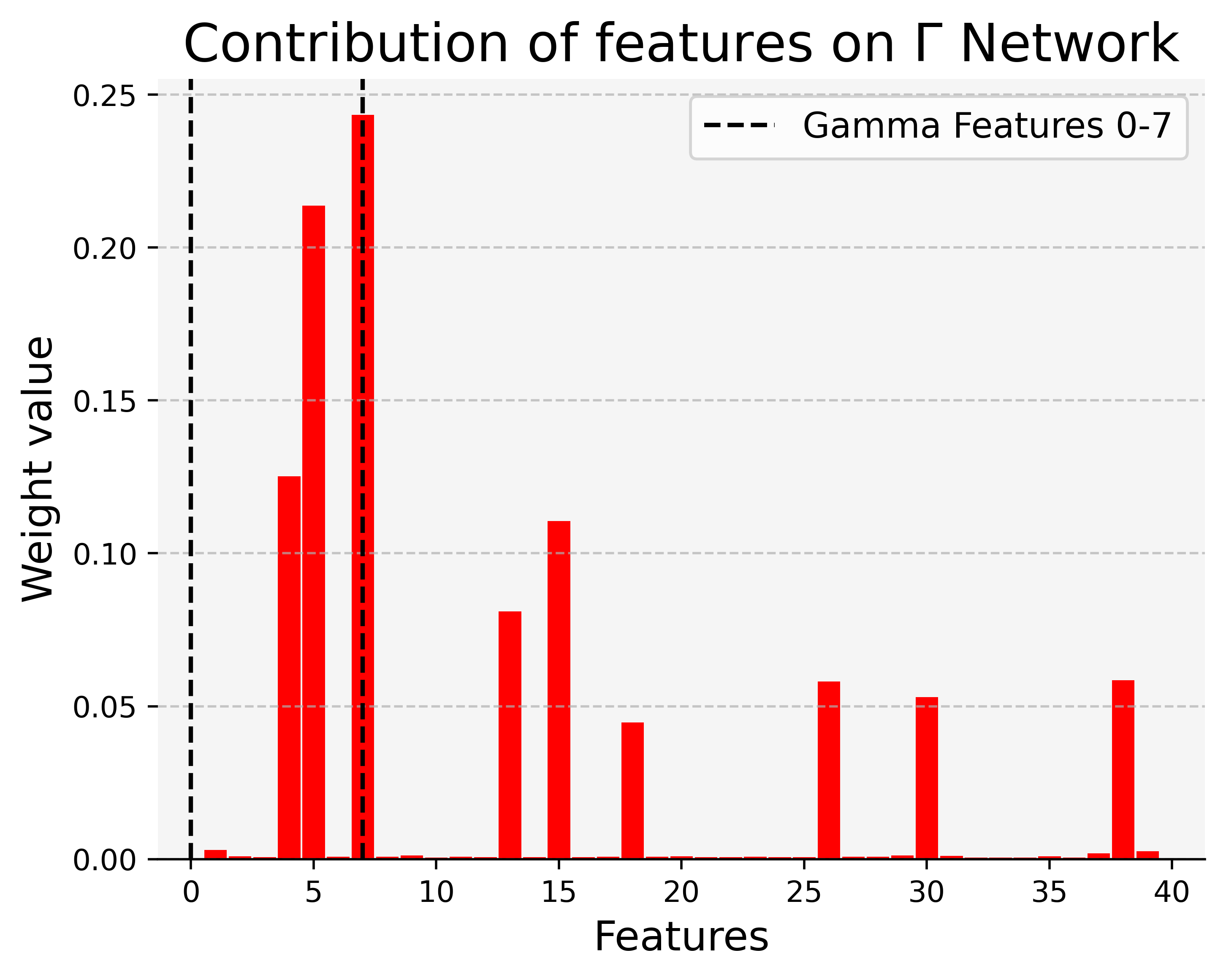}
			\includegraphics[width=0.22\textwidth]{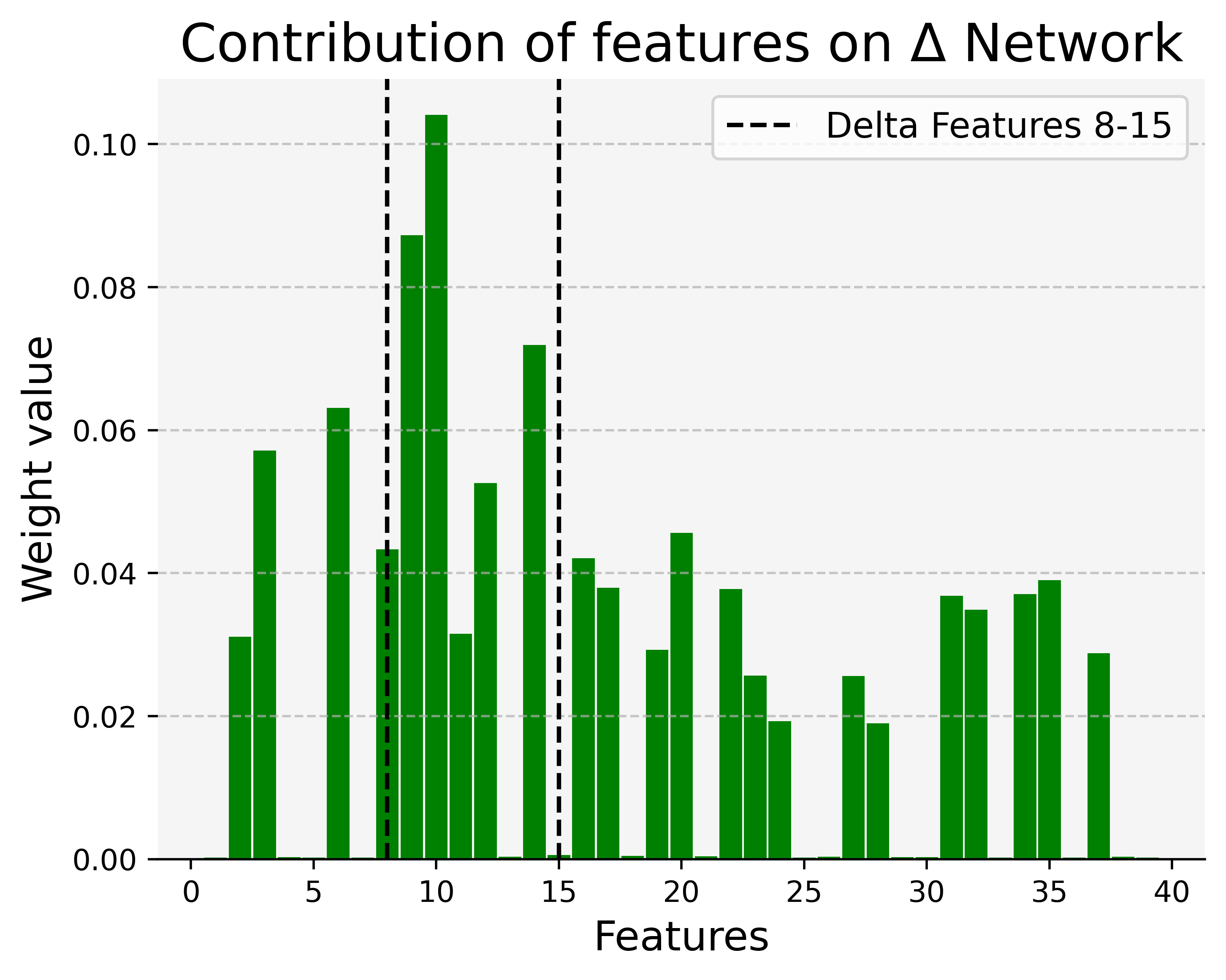}
			\includegraphics[width=0.22\textwidth]{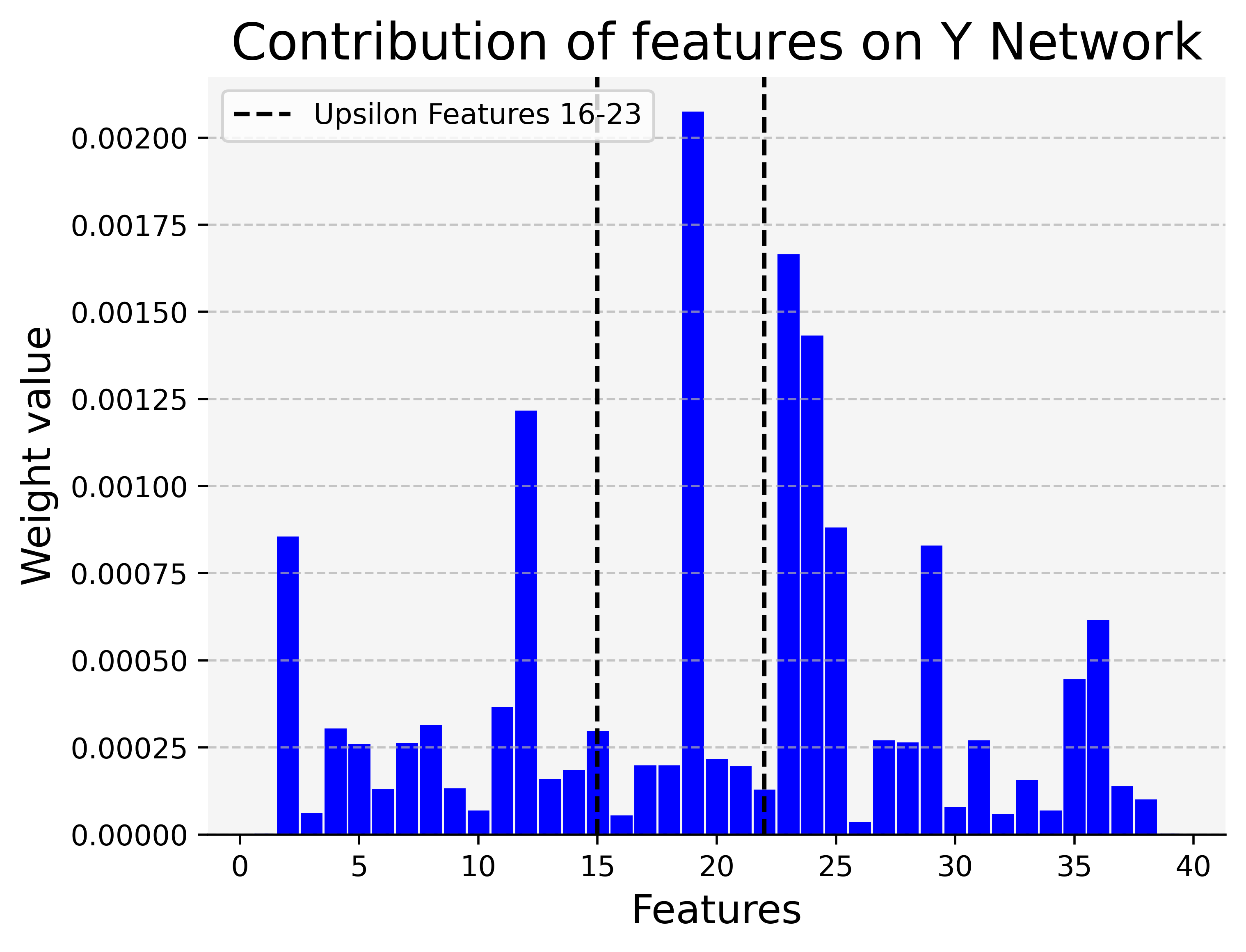}
			\includegraphics[width=0.22\textwidth]{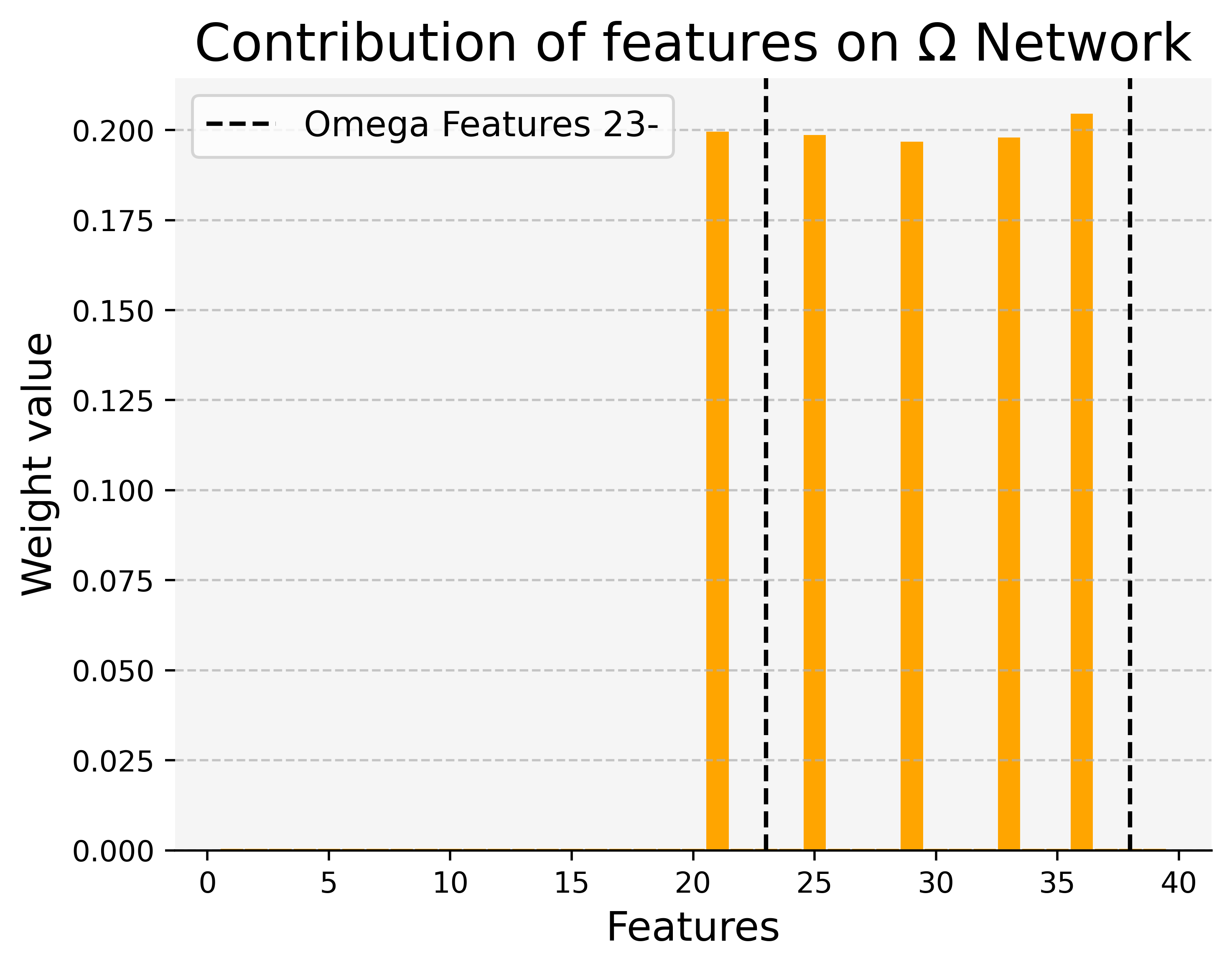}
			\includegraphics[width=0.22\textwidth]{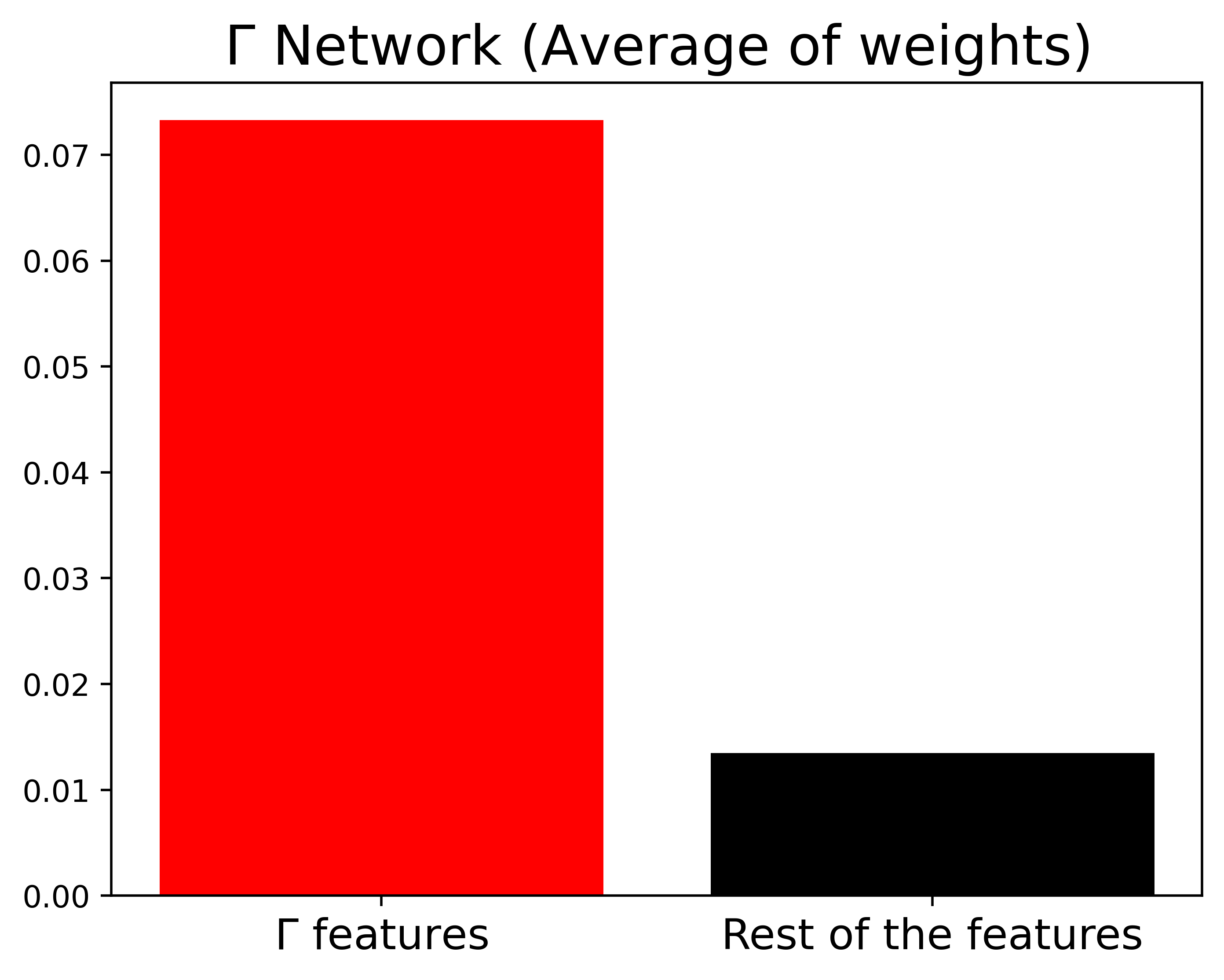}
			\includegraphics[width=0.22\textwidth]{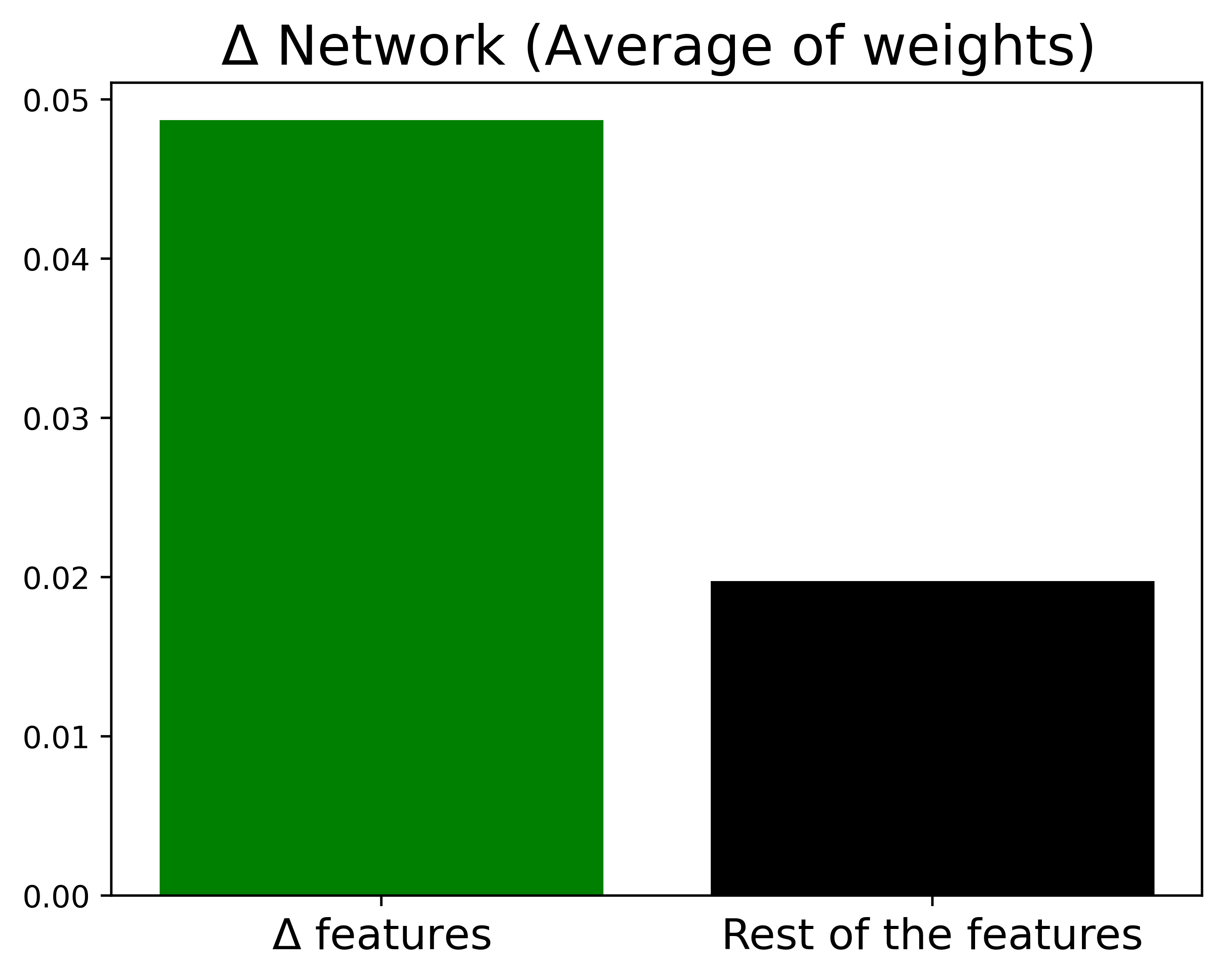}
			\includegraphics[width=0.22\textwidth]{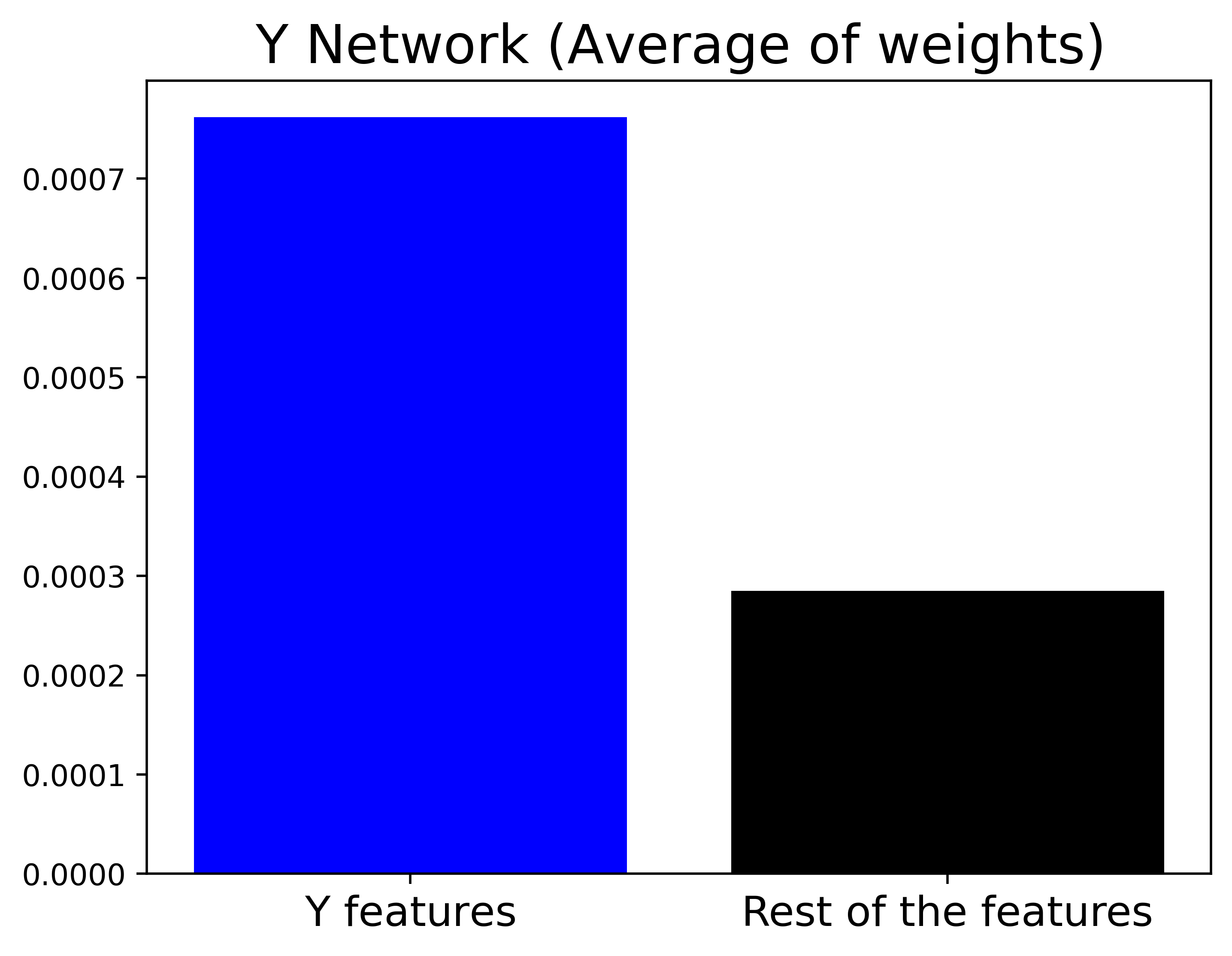}
			\includegraphics[width=0.22\textwidth]{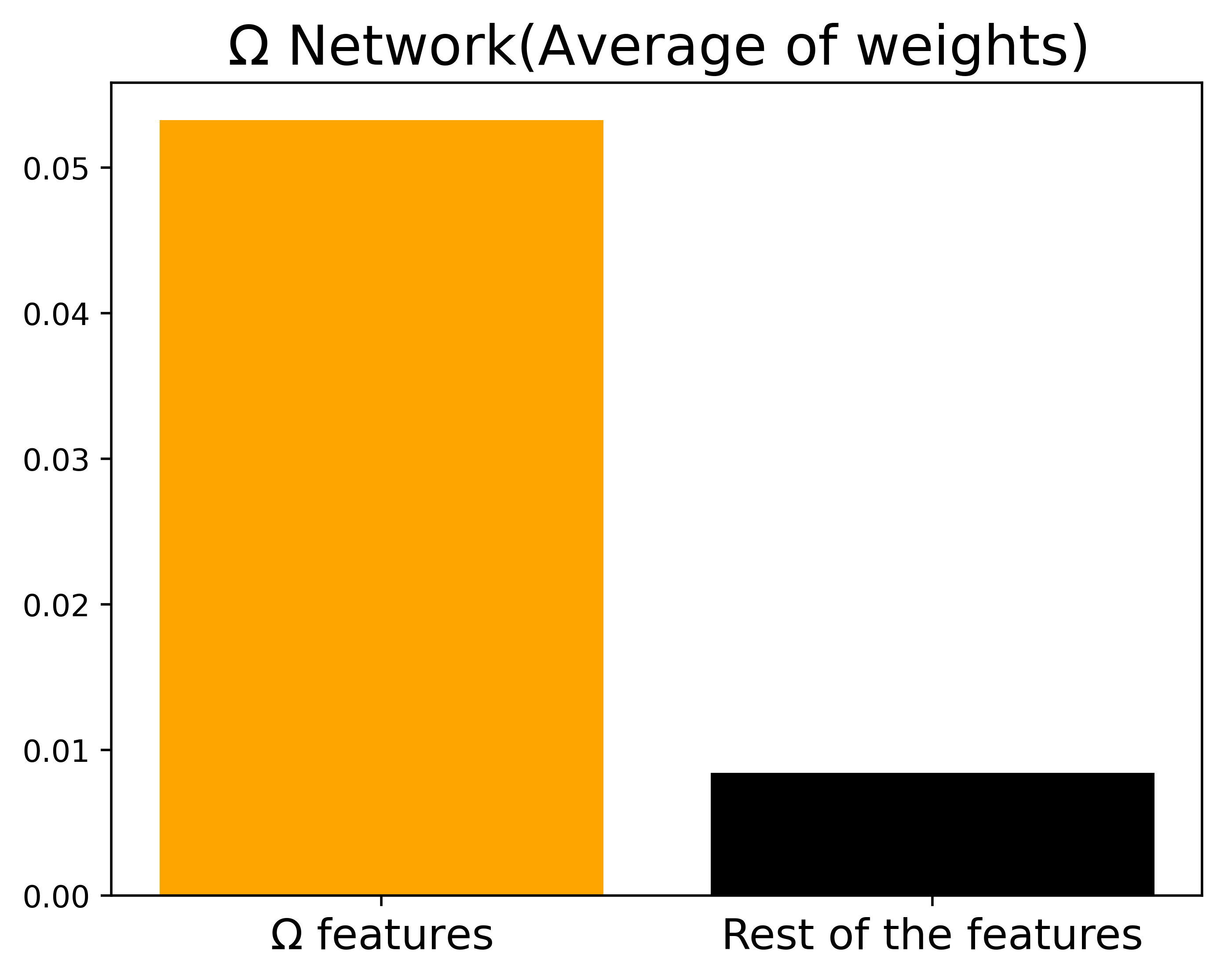}
		
		\caption{The visualization of feature contributions on each latent factor representational network is conducted for the dataset with dimensions 8, 8, 8, 15 ($\Gamma$,$\Delta$,$\Upsilon$,$\Omega$) utilizing the $\bar{W}$ criterion based on DRI-ITE (ours). The top row visualizes all individual features, where high values are expected for the features between dotted lines, the bottom row represents the average over all features that are supposed to be represented by that particular network compared to the average weight of wrongly represented features.}
		
		\label{fig:ident_all}
	\end{figure*}

\begin{figure*}
	\centering
	
		\includegraphics[width=0.28\textwidth]{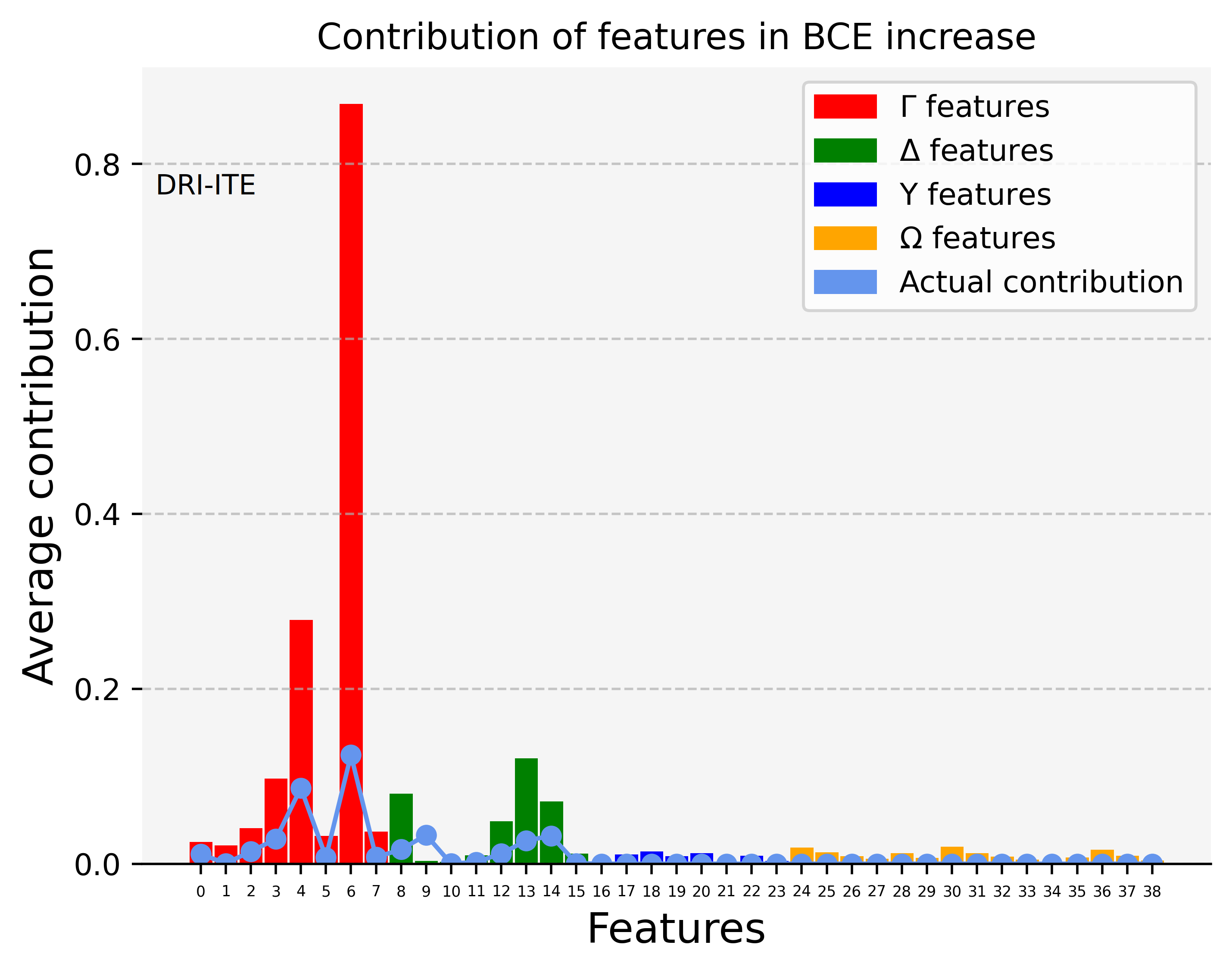}
		\includegraphics[width=0.28\textwidth]{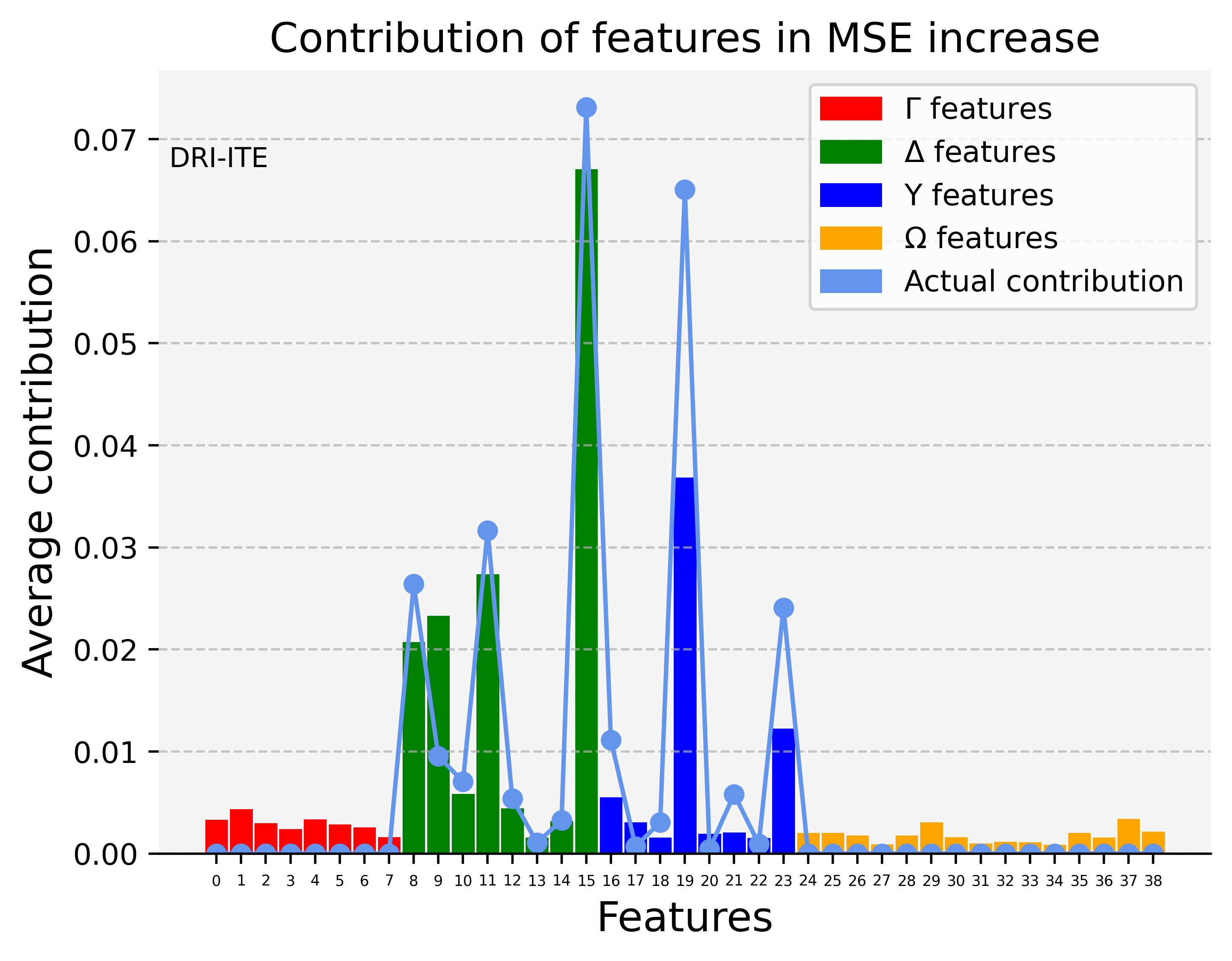}
		\includegraphics[width=0.28\textwidth]{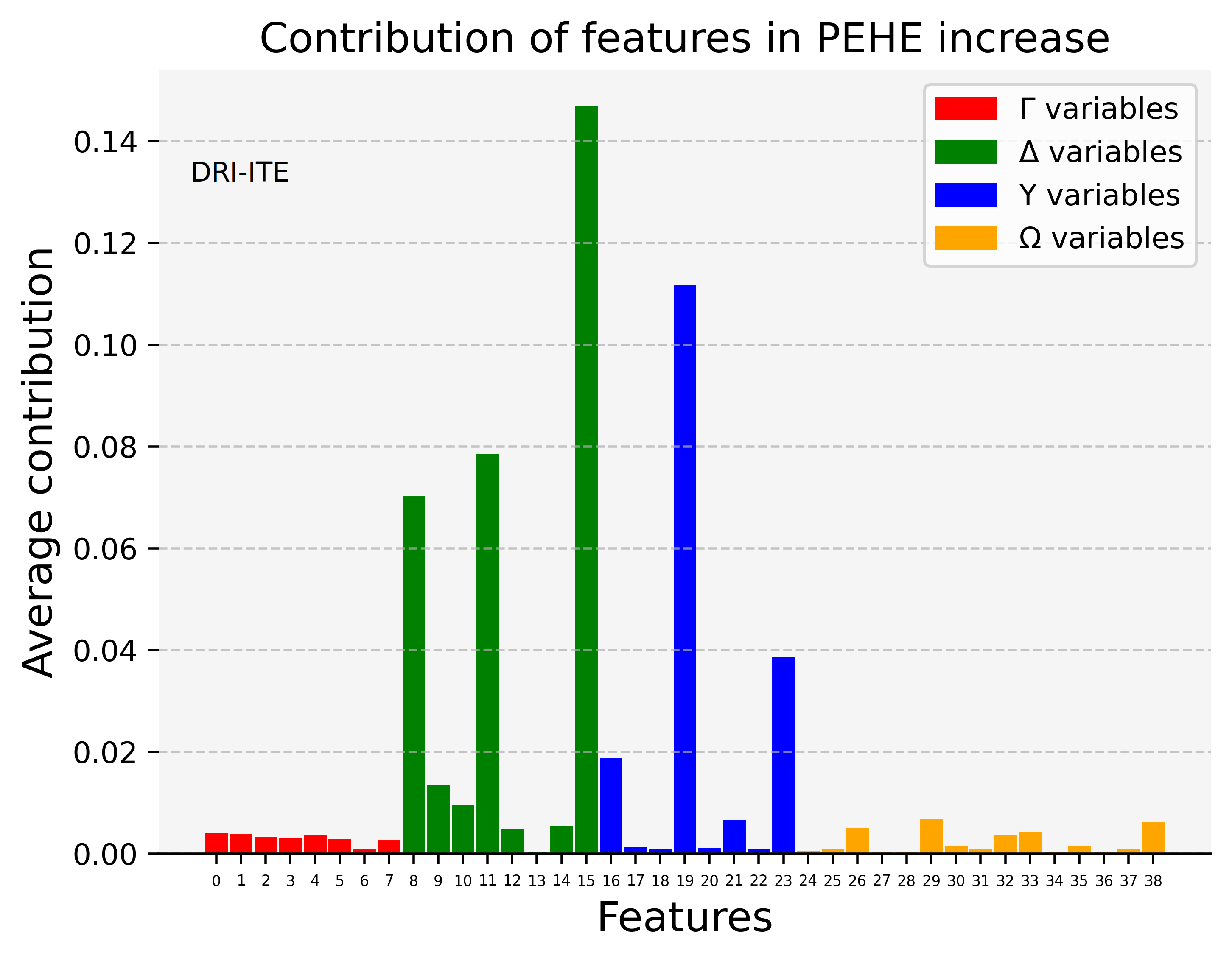}
	
	\caption{The average contribution of each feature in increasing BCE, MSE and PEHE loss is assessed by permuting features based on DRI-ITE (ours). Figure (a) shows the average contribution of each feature in increasing BCE (contribution of $\Gamma$ and $\Delta$ features should have higher increases in BCE as compared to rest of the features, if $\Gamma$ and $\Delta$ factors are identified correctly). Figure (b) shows the average contribution of each feature in increasing MSE (contribution of $\Delta$ and $\Upsilon$ features should have higher increases in MSE, if $\Delta$ and $\Upsilon$ factors are identified correctly). Figure(c) shows the average increase in PEHE (0.0017) by irrelevant variables using DRI-ITE (ours). A lower impact of irrelevant variables in increasing PEHE indicates accurate disentanglement of $\Omega$ and reliable ITE estimation.}
	
	\label{fig:ident_BCE_MSE}
\end{figure*}

\begin{figure*}
	\centering

		\includegraphics[width=0.26\textwidth]{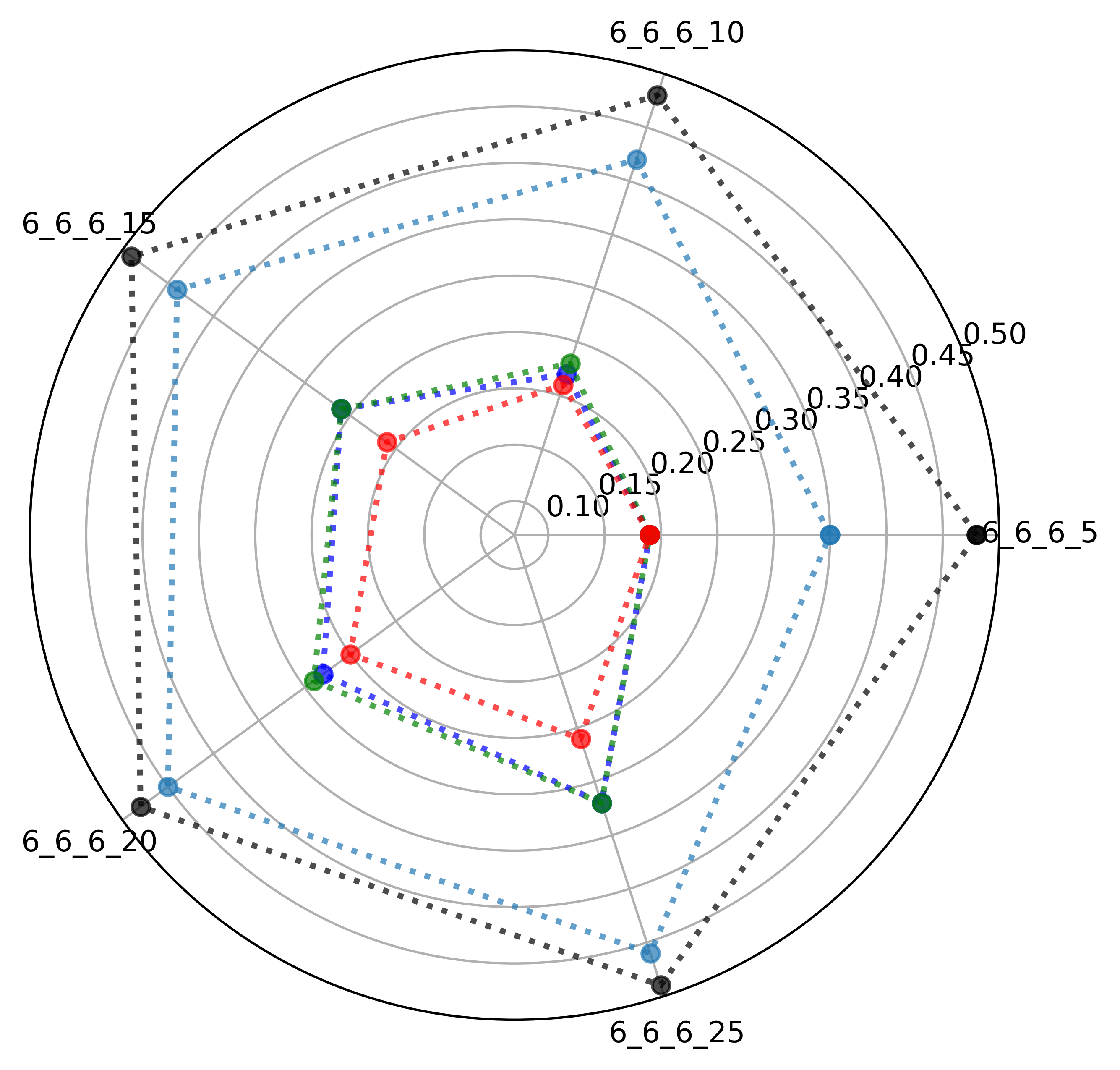}\hskip 0.7cm
		\includegraphics[width=0.29\textwidth]{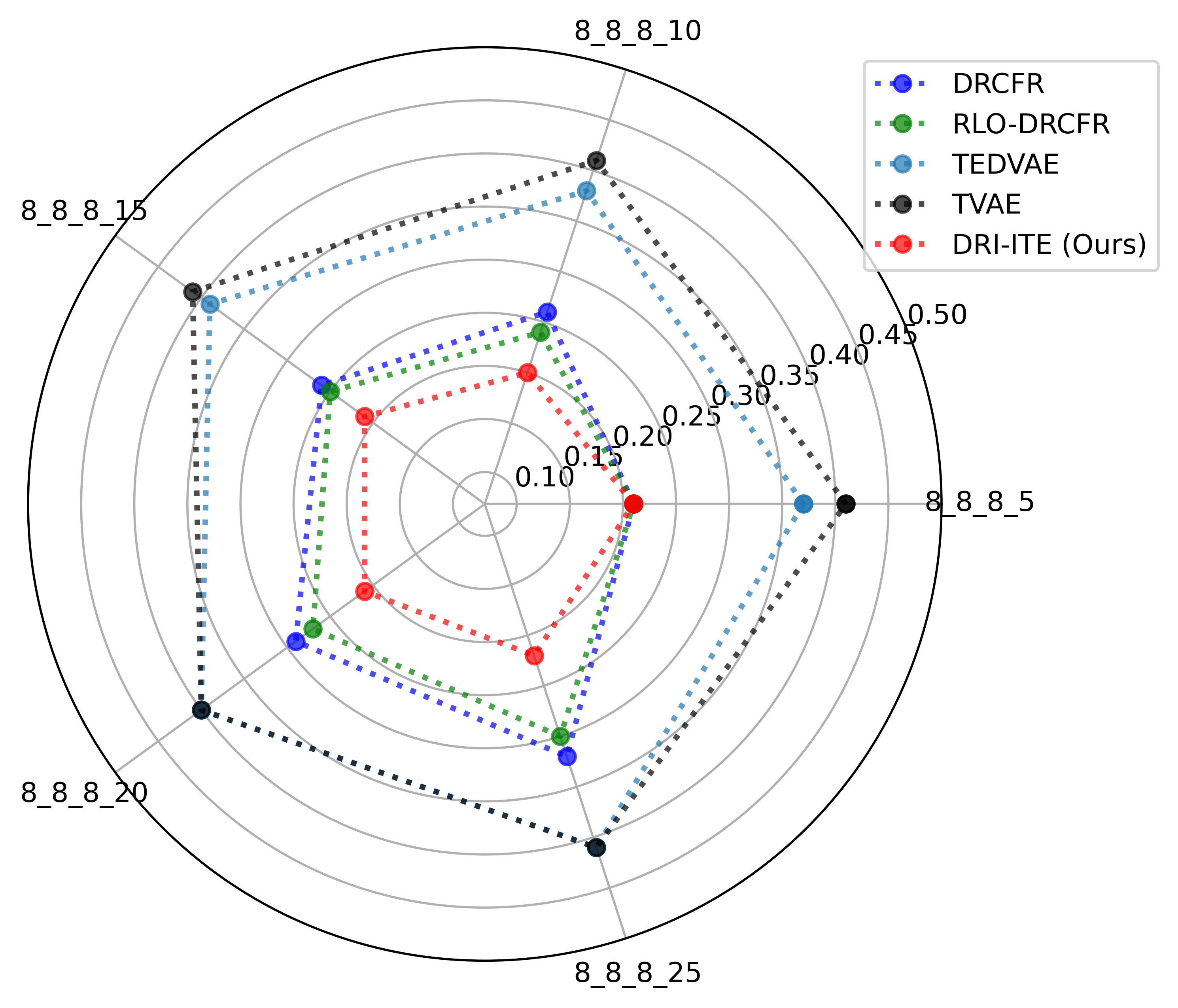}
		\includegraphics[width=0.268\textwidth]{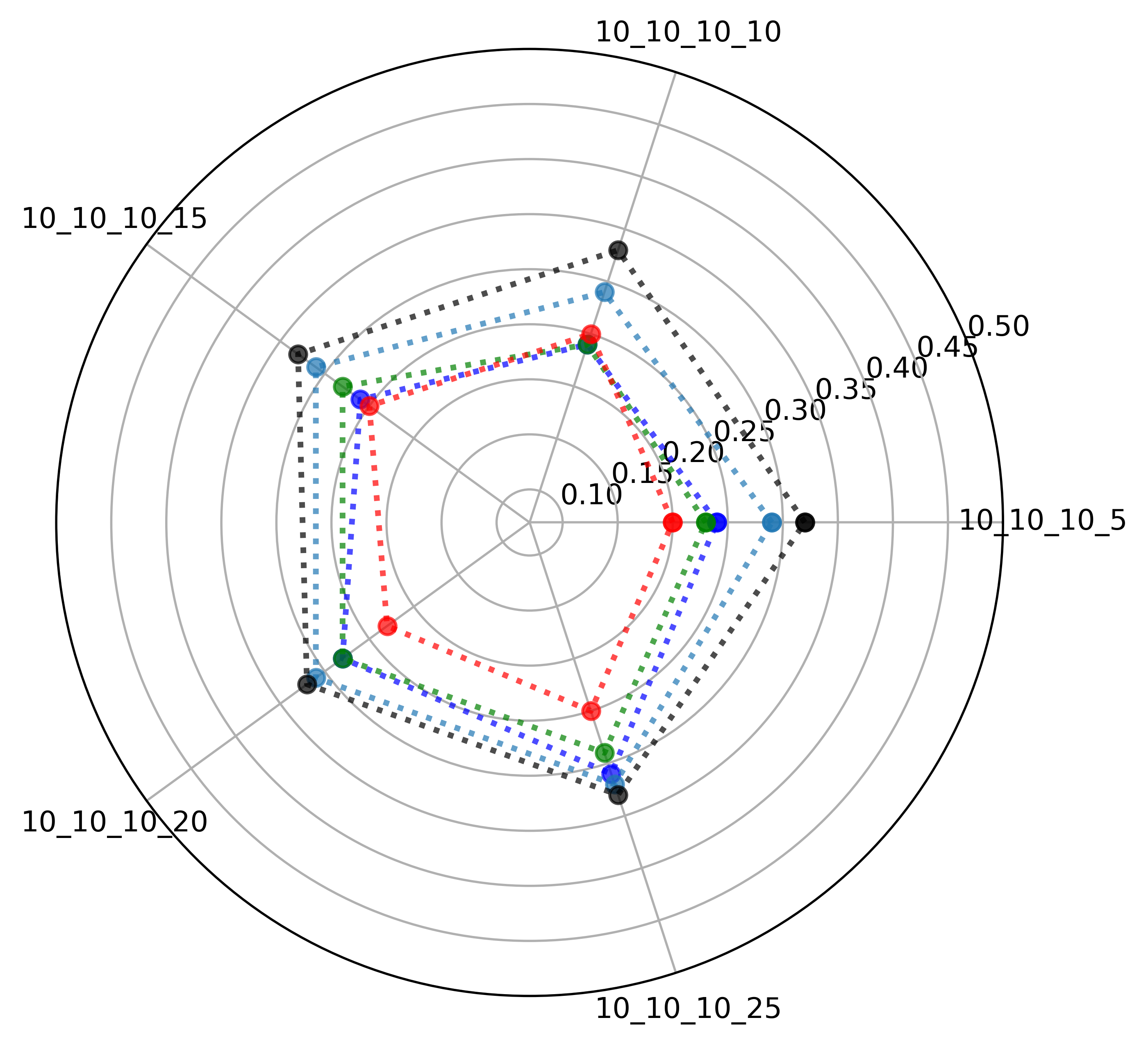}

	\caption{Radar charts visualizing the PEHE (mean values) results on the synthetic dataset. Each vertex represents the dimensions of latent factors ($\Gamma$\_$\Delta$\_$\Upsilon$\_$\Omega$). PEHE values closer to the center are better. Dashed red lines show our proposed method (DRI-ITE).}
	\label{fig:syn_pehe}
\end{figure*}

\begin{table*}[t]
	
	\small
	
	\centering
	
	\caption{Results of ablation study of loss function(bold numbers indicate smallest/best results).}
	\addtolength{\tabcolsep}{-1pt}
	\begin{tabular}{ l|lllll}  
		
		\toprule
		
		{ Loss}  & {8$\_$8$\_$8$\_$5}  & {8$\_$8$\_$8$\_$10} & {8$\_$8$\_$8$\_$15} & {8$\_$8$\_$8$\_$20} & {8$\_$8$\_$8$\_$25}\\
		\midrule
		
		$\mathcal{L}_{reg}+\mathcal{L}_{class}+\mathcal{L}_{disc}$ & \textbf{0.21{\tiny(0.009)}}&  0.25{\tiny(0.01)} & {0.26{\tiny(0.01)}} &  {0.28{\tiny(0.01)}} & {0.32{\tiny(0.006)}}
		\\
		$\mathcal{L}_{reg}+\mathcal{L}_{class}+\mathcal{L}_{disc}+\mathcal{L}_{orth}$  & 0.21{\tiny(0.01)}&  0.25{\tiny(0.02)} & {0.25{\tiny(0.01)}} & {0.28{\tiny(0.01)}} & {0.31{\tiny(0.005)}}
		\\
		$\mathcal{L}_{reg}+\mathcal{L}_{class}+\mathcal{L}_{disc}+\mathcal{L}_{orth}+\mathcal{L}_{recons}$  & 0.22{\tiny(0.01)} & \textbf{0.20{\tiny(0.01)}} & \textbf{0.20{\tiny(0.02)}} & \textbf{0.21{\tiny(0.01)}} & \textbf{0.21{\tiny(0.01)}}
		\\

		\bottomrule
	\end{tabular}
	
	\label{tab:ablation}
\end{table*}

\section{Experiments}
\label{sec:experiments}
Before discussing the results of the proposed method, we will briefly discuss the used datasets, evaluation criteria and experiment details. Our code is available \citep{Khan}.


\subsection{Datasets}

We use both synthetic and real-world datasets to evaluate the performance of the proposed method. A synthetic dataset allows to control all the latent factor that make up $\mathcal{X}$. To this purpose, we are augmenting the existing dataset proposed by \citet{Negar} with additional irrelevant variables. Additionally, performance on the commonly used IHDP and jobs dataset will be analyzed.

\subsubsection{Synthetic Dataset}

The dataset comprises a sample size of $N$, with dimensions $\left[{m}_{\Gamma},{m}_{\Delta},{m}_{\Upsilon} \right]$, along with mean and covariance matrices ($\mu_{L},\sum_{L}$) for each latent factor $L \in \left[\Gamma,\Delta,\Upsilon\right]$. A multivariate normal distribution is employed for data generation, and the covariates matrix is constructed as $N \times (m_{\Gamma}+m_{\Delta}+m_{\Upsilon})$. The synthetic dataset is generated using the same settings and approach as presented by \citet{Negar}. To generate $\Omega$, we follow the feature selection community by adding artificial contrasts. Each irrelevant variable is a permutation of a randomly selected feature generated for the other factors as simply using Gaussian or uniform distributions may not be sufficient \citep{Irre}.


\subsubsection{Infant Health and Development Program (IHDP)}

IHDP is a binary treatment dataset based on experiment conducted by \citet{Gunn}. \citet{Hill} introduced selection bias in original RCT data to make it an observational dataset. It contains 25 covariates that describe different aspects of the child and mother, such as birth weight, neonatal health index, mother's age, drug status, etc. The data has 747 instances in total, 139 belong to the treated group and 608 belong to the control group. The purpose of the study/data was to check the effect of treatment (specialist home visits) on the cognitive health of children. IHDP does not contain irrelevant variables, therefore we augment it with artificial contrasts for the evaluation purpose.

\subsubsection{Jobs}

Jobs is an observational dataset collected under the Lalonde experiment \citep{lalonde}. It contains eight pre-treatment covariates; age, educ, black, hisp, married, nodegr, re74, re75. The treatment data is binary and shows whether a person received job training or not. At the same time, the outcome variable indicates the earnings of a person in 1978. The data has 614 instances in total, 185 belong to the treated group, and 429 belong to the control group \citep{Hill,jobs}. We use artificial contrasts for the evaluation purpose.



\subsection{Evaluation criteria}
The well-established criterion for treatment effect estimation is Precision in Estimation of Heterogeneous Effect (PEHE), which is defined as follows:
\begin{eqnarray}
	\mathit{PEHE}=\sqrt{\frac{1}{N}\sum_{i=1}^{N}(\hat{e}_{i}-e_{i})^2}
\end{eqnarray}

where $\hat{e}_{i}=\hat{y}_{i}^{1}-\hat{y}_{i}^{0}$ and ${e}_{i}={y}_{i}^{1}-{y}_{i}^{0}$ are predict and true effects respectively. 

Secondly, we use another well-known criterion, Policy Risk ($\mathcal{R}_{pol}$), which is defined as under:

\begin{eqnarray}
	\begin{aligned}
		\mathcal{R}_{{pol}}(\pi_f) = & 1 - ( \E[Y^1 \,|\, \pi_f(x) = 1] \cdot p(\pi_f = 1) \\& + \E[Y^0 \,|\, \pi_f(x) = 0] \cdot p(\pi_f = 0))
	\end{aligned}
\end{eqnarray}

The policy risk is a measure of the average loss in value when following a specific treatment policy. The treatment policy ($\pi_f(x)$) is a set of rules based on the predictions of a model $f$. Specifically, if the difference in the model's predictions for treatment (1) and no treatment (0) is greater than a threshold ($\lambda$), then treat ($\pi_f(x) = 1$), otherwise do not treat ($\pi_f(x) = 0$). This formula involves the expected outcomes when following the treatment policy, weighted by the probabilities of applying the policy \citep{UriSha}.
\subsection{Experiment details}

\begin{table}[h]
	
	\small
	
	\centering
	
	\caption{Hyper-parameters and Ranges.}
	\addtolength{\tabcolsep}{-1pt}
	\begin{tabular}{ l|l}  
		
		\toprule
		
		{Hyper-parameter}  & {Range} \\
		\midrule
		
		{Latent dimensions}  & {\{5, 10, 15, 100\}} \\
		
		{Hidden dimensions}  & {\{50, 100, 200\}} \\
		
		{Layers}  & {\{2, 3, 4\}} \\
		{Batch size}  & {\{32, 64, 128, 256\}} \\
		{Learning rate}  & {\{$1{e}{-2}$, $1{e}{-3}$, $1{e}{-4}$, $1{e}{-5}$\}} \\
		{$\alpha$, $\beta$, $\gamma$, $\lambda$, $\mu$}  & {\{0.01, 0.1, 1, 5, 10, 100\}} \\
		\bottomrule
	\end{tabular}
	
	\label{tab:hyper}
\end{table}

We employed three layers for the representational network of each latent factor ($\Gamma$, $\Delta$, $\Upsilon$, $\Omega$). The hidden and output layer for $\Gamma$, $\Delta$, $\Upsilon, \Omega$ consisted of $10, 15$ neurons across multiple experiments. We utilized Adam as the optimizer, and ELU served as the activation function. The batch size was set to 256, the number of epochs to 5000 maximum, and the learning rate to $1{e}{-5}$. Following the approach outlined in \citep{UriSha}, we employed $PEHE_{nn}$ on the validation set to save the best model. The data split between training and testing mirrored that used in \citep{UriSha,Negar}, with $20\%$ of the training data reserved for the validation set. We used the same settings for jobs and synthetic datasets but only employed 100 dimensional representational networks to assign enough capacity for fair comparisons. 

To select hyper-parameters, we employed grid search across different ranges (see Table \ref{tab:hyper}). These parameter ranges were inspired by various baseline methods. 



\subsection{Results}

We evaluate our method on two evaluation criteria: how accurately it identifies all disentangled latent factors  (Subsection \Ref{sec:Identification}) and secondly how effectively it estimates the treatment effect using PEHE and policy risk ($\mathcal{R}_{pol}$) criterion (Subsection \Ref{sec:Evaluation}).

\begin{table*}[t]

	\centering
	
	\caption{PEHE (mean (std)) on IHDP with different dimensions of $\Omega$ and varied latent dimensions of representational networks (bold numbers indicate smallest/best results).}
	\addtolength{\tabcolsep}{-3.5pt}
	\begin{tabular}{l|lllll|lllll}  
		
		\toprule
		
		&\multicolumn{5}{c}{Latent dimensions=10} & \multicolumn{5}{c}{Latent dimensions=15}\\ 
		
		{ Data}$\_\Omega$  & { DR-CFR}  & { RLO-DRCFR} & { TEDVAE} & { TVAE}& { DRI-ITE(Ours)}& { DR-CFR}  & { RLO-DRCFR} & { TEDVAE} & { TVAE} & { DRI-ITE(Ours)}\\
		\midrule

		IHDP$\_$5 
		& 1.30{\tiny(0.78)}& 1.33{\tiny(0.81)} & \textbf{0.95{\tiny(0.62)}} &{1.25{\tiny(0.38)}}& {1.12{\tiny(0.62)}}
		& 1.19{\tiny(0.62)}& 1.26{\tiny(0.71)} & \textbf{0.93{\tiny(0.62)}} &{1.28{\tiny(0.56)}}& {1.06{\tiny(0.60)}} \\
		
		IHDP$\_$10  
		& 1.48{\tiny(0.94)}& 1.36{\tiny(0.76)} & {1.18{\tiny(0.80)}} &{1.29{\tiny(0.43)}}& \textbf{1.12{\tiny(0.65)}} 
		& 1.25{\tiny(0.73)}& 1.34{\tiny(0.72)} & \textbf{1.15{\tiny(0.82)}} &{1.31{\tiny(0.46)}}& {1.20{\tiny(0.66)}}\\
		
		IHDP$\_$15 
		& 1.51{\tiny(0.98)}& 1.37{\tiny(0.78)} & {1.33{\tiny(0.83)}} &{1.43{\tiny(0.58)}}& \textbf{1.21{\tiny(0.69)}} 
		& 1.29{\tiny(0.73)}& 1.36{\tiny(0.77)} & {1.35{\tiny(0.86)}} &{1.29{\tiny(0.51)}}& \textbf{1.23{\tiny(0.65)}}\\
		
		IHDP$\_$20  
		& 1.52{\tiny(1.01)}& 1.49{\tiny(0.91)} & {1.42{\tiny(0.94)}} &\textbf{1.23{\tiny(0.48)}}& \textbf{1.23{\tiny(0.65)}}
		& 1.30{\tiny(0.74)}& 1.30{\tiny(0.70)} & {1.41{\tiny(0.89)}} &{1.23{\tiny(0.48)}}& \textbf{1.15{\tiny(0.61)}}\\

		\bottomrule
	\end{tabular}
	
	\label{tab:pehe_ihdp}
\end{table*}

\subsubsection{Identification of Disentangled Latent Factors} \label{sec:Identification}

To quantify, how precisely our proposed method identifies all latent factors in the disentangled embedding spaces we use the following metrics:    
\begin{itemize}
	\item \textbf{Calculation of average weight vector: }We compute $W$, defined as the product of weight matrices across all layers within a representational network, and $\bar{W}$, which represents the row-wise average vector of the absolute values of $W$ for each representational network. The vector $\bar{W}$ provides insight into the average contribution of each feature within that specific representational network.
	
	In the case of synthetic data, where the assignment of features to latent factors is known, we generated post-training plots of $\bar{W}$ for each network. The rationale behind this analysis \citep{Anpeng} lies in the expectation that the average weights corresponding to features associated with a particular latent factor should exhibit higher values compared to other features.
	
	\item \textbf{Permutation feature importance analysis: }Secondly, we employed permutation feature importance theory \citep{Fisher} to validate the precise disentanglement of latent factors achieved by our representational networks. The underlying principle is straightforward: if shuffling a feature leads to an increase in model error after training, the feature is deemed important; otherwise, it can be considered unimportant. To the best of our knowledge, this is the first attempt to apply permutation feature importance theory to the domain of treatment effect estimation
	
\end{itemize}


In Figure \ref{fig:ident_all}, the top half shows the $\bar{W}$ bar plots for the $\Gamma$, $\Delta$, $\Upsilon$, and $\Omega$ representational networks using synthetic data. Notably, only relevant features exhibit high weights compared to the remaining ones. The figure confirms that each network accurately identifies its corresponding latent factors while effectively avoiding information leakage among them. The bottom half of Figure \ref{fig:ident_all} presents the average weights of the respective features (between vertical lines) and the remaining features for each representational network. This visualization emphasizes that our approach selectively focuses on relevant information for each network, leading to accurate identification of latent factors.


Figure \ref{fig:ident_BCE_MSE} illustrates the identification of latent factors based on the second criterion of permutation feature importance theory \citep{Fisher}. Specifically, Figure \ref{fig:ident_BCE_MSE} (a) vividly demonstrates that only $\Gamma$ and $\Delta$ features actively contribute to increasing BCE loss. This observation supports the conclusion that our method accurately identifies $\Gamma$ and $\Delta$ factors from the data. Likewise, Figure \ref{fig:ident_BCE_MSE} (b) confirms the successful identification of $\Delta$ and $\Upsilon$ factors indicated by the increased MSE. 

The illustration in Figure \ref{fig:ident_BCE_MSE} (c) shows again that DRI-ITE accurately disentangles and identifies $\Omega$, as permuting irrelevant features does not increase the PEHE, while permuting any relevant feature does. We conjecture that the baseline methods fail to capture the feature importance completely. If true, this should result in overall lower ITE estimation errors.

Investigating the synthetic dataset, it is evident from Figure \ref{fig:syn_pehe} that DRI-ITE consistently outperforms the baseline methods on PEHE evaluation. As the dimensions of $\Omega$ increase, baseline methods experience a much stronger decline in performance. DRI-ITE demonstrates better performance, particularly in scenarios with high-dimensional $\Omega$.

In Table \ref{tab:ablation}, we perform an ablation study to analyze the impact of the components $\mathcal{L}_{orth}$ and $\mathcal{L}_{recons}$ on the PEHE compared to the basic loss (i.e., $\mathcal{L}_{reg}+\mathcal{L}_{class}+\mathcal{L}_{disc}$). The results show that the addition of the orthogonal loss results in a minor improvement in performance, while the addition of the reconstruction loss leads to a significant decrease in the PEHE when ten or more irrelevant variables are introduced to the original variable set.

\subsubsection{Evaluation on Estimation of Treatment Effect} \label{sec:Evaluation}
Successfully disentangling latent factors including $\Omega$ in itself is not enough, but ultimately we aim to have improved estimates of the ITE. We assessed the performance of DRI-ITE using PEHE on the IHDP benchmark dataset; and using policy risk ($\mathcal{R}_{pol}$) criterion on Jobs dataset.

We are comparing our results with four SOTA baseline disentanglement approaches.
\begin{itemize}
	\item Disentangled Representations for Counterfactual Regression: \textbf{DR-CFR} \citep{Negar}.
	\item Learning Disentangled Representations for Counterfactual Regression via Mutual Information Minimization: \textbf{RLO-DRCFR} \citep{Ortho}.
	\item Treatment Effect with Disentangled Autoencoder: \textbf {TEDVAE} \citep{TEDEV}.
	\item Targeted VAE: Variational and Targeted Learning
	for Causal Inference: \textbf{TVAE} \citep{vowels2021targeted}
	
\end{itemize}

\begin{table}[h]

	\centering

	\caption{Policy risk (mean (std)) on Jobs with different dimensions of $\Omega$ (bold numbers indicate smallest/best results).}
	\addtolength{\tabcolsep}{-1pt}
	\begin{tabular}{ llllll}  
		
		\toprule
		
		
		{\fontsize{7}{8}\selectfont Data}$\_\Omega$  & {\fontsize{7}{8}\selectfont DR-CFR}  & {\fontsize{7}{8}\selectfont RLO-DRCFR} & {\fontsize{7}{8}\selectfont TEDVAE} & {\fontsize{7}{8}\selectfont TVAE} & {\fontsize{7}{8}\selectfont DRI-ITE(Ours)} \\
		\midrule

		Jobs$\_$5 & 0.13{\tiny(0.03)}&  0.13{\tiny(0.03)} & {0.20{\tiny(0.03)}}  & {0.14{\tiny(0.01)}} & \textbf{0.11{\tiny(0.02)}} 
		\\
		
		
		Jobs$\_$15  & \textbf{0.12}{\tiny(0.03)} & \textbf{0.12}{\tiny(0.03)} & {0.21{\tiny(0.04)}} & {0.15{\tiny(0.01)}} & \textbf{0.12{\tiny(0.04)}} 
		\\
		
		Jobs$\_$20 & 0.14{\tiny(0.04)} & 0.13{\tiny(0.04)} & {0.19{\tiny(0.03)}} & {0.22{\tiny(0.08)}} & \textbf{0.11{\tiny(0.02)}} \\
		
		\bottomrule
	\end{tabular}
	
	\label{tab:rp_jobs}
\end{table}

We evaluated DRI-ITE on IHDP. Table \ref{tab:pehe_ihdp} presents PEHE values on the widely used IHDP benchmark dataset. The PEHE values (mean{\tiny(std)}) are calculated from the first $30$ realizations of IHDP, incorporating different dimensions of $\Omega$ and varying latent dimensions of representational networks.

Again, the performance of SOTA methods tends to degrade strongly with increasing dimensions of $\Omega$. As depicted in Table \ref{tab:pehe_ihdp}, DRI-ITE effectively maintains a low PEHE in comparison to baseline methods after the introduction of $\Omega$. Particularly noteworthy is the struggle of baseline methods, in scenarios with low-dimensional representational networks, which supposedly suppress $\Omega$ through regularization. This results in the assimilation of information from $\Omega$ into other relevant factors, consequently leading to poor performance. In contrast, our method adeptly disentangles $\Omega$ and consistently outperforms baseline methods.

However, as the representational networks increase in dimensionality, the performance of baseline methods also improves. We observed that for the baselines, regularization becomes more effective in suppressing $\Omega$ in high-dimensional scenarios compared to low dimensional networks. Despite this, our approach continues to provide better results. However, TEDVAE shows good performance against small number of $\Omega$ but it fails to ignore $\Omega$ with higher dimensions. 

Table \ref{tab:rp_jobs} presents a comparison between baseline methods and DRI-ITE regarding policy risk ($\mathcal{R}_{pol}$) criteria. These results are estimates (mean(std)) derived from the initial 30 realizations of the jobs dataset. Notably, the table illustrates that the performance of DRI-ITE remains consistently better and unaffected by the inclusion of $\Omega$. In contrast, baseline methods experience a decline in performance as the dimensionality of $\Omega$ increases.


These results substantiate our assertion that SOTA methods lack an explicit and reliable mechanism to disentangle or ignore $\Omega$. Conversely, our approach consistently disentangles $\Omega$ factors and reliably estimates ITE across all scenarios in comparison to SOTA methods. Moreover, these results are statistically significant based on the t-test with $\alpha=0.05$.

\section{Conclusion}
\label{sec:conclusion}



In this paper, we address the problem of learning disentangled representation for Individual Treatment Effect (ITE) estimation with observational data. While deep disentanglement-based methods have been widely employed, they face limitations in handling irrelevant factors, leading to prediction errors. In the era of data-driven and big data approaches, where pre-screening for relevance is impractical, our work seeks to provide a robust solution to the inevitable presence of irrelevant factors in observational studies. We present a novel approach that goes beyond traditional deep disentanglement methods by explicitly identifying and representing irrelevant factors, in addition to instrumental, confounding, and adjustment factors. Our method leverages a deep embedding technique, introducing a reconstruction objective to create a dedicated embedding space for irrelevant factors through an autoencoder. Our empirical experiments, conducted on synthetic and real-world benchmark datasets, demonstrate the efficacy of our method. We showcase an improved ability to identify irrelevant factors and achieve more precise predictions of treatment effects compared to previous approaches. While our approach primarily addresses the scenario with two treatment groups, in future we plan to work with multiple and continuous treatments.

\begin{ack}
This work has been supported by the Industrial Graduate School Collaborative AI \& Robotics funded by the Swedish Knowledge Foundation Dnr:20190128, and the Knut and Alice Wallenberg Foundation through Wallenberg AI, Autonomous Systems and Software Program (WASP).
\end{ack}
\vspace{2cm}


\bibliography{mybibfile}

\end{document}